\documentclass{article}

\usepackage{microtype}
\usepackage{graphicx}
\usepackage{subfigure}
\usepackage{booktabs} 

\usepackage{hyperref}

\usepackage[accepted]{icml2024}

\usepackage{amsmath}
\usepackage{amssymb}
\usepackage{mathtools}
\usepackage{amsthm}

\usepackage[capitalize,noabbrev]{cleveref}

\theoremstyle{plain}

\theoremstyle{definition}

\theoremstyle{remark}

\usepackage[textsize=tiny]{todonotes}

\icmltitlerunning{Unifying Image Processing as Visual Prompting Question Answering}

\begin{document}
	
	\twocolumn[
	\icmltitle{Unifying Image Processing as Visual Prompting Question Answering}
	
	
	
	\icmlsetsymbol{equal}{*}
	
	\begin{icmlauthorlist}
		\icmlauthor{Yihao Liu}{equal,ailab,siat}
		\icmlauthor{Xiangyu Chen}{equal,ailab,siat,um}
		\icmlauthor{Xianzheng Ma}{equal,ailab}
		\icmlauthor{Xintao Wang}{arc}
		\icmlauthor{Jiantao Zhou}{um}
		\icmlauthor{Yu Qiao}{ailab,siat}
		\icmlauthor{Chao Dong}{siat,ailab}
		
	\end{icmlauthorlist}
	
	\icmlaffiliation{ailab}{Shanghai Artificial Intelligence Laboratory}
	\icmlaffiliation{siat}{Shenzhen Institute of Advanced Technology, Chinese Academy of Sciences}
	\icmlaffiliation{um}{University of Macau}
	\icmlaffiliation{arc}{ARC Lab, Tencent PCG}
	
	\icmlcorrespondingauthor{Chao Dong}{chao.dong@siat.ac.cn}
	
	\icmlkeywords{Machine Learning, ICML}
	
	\vskip 0.3in
	]
	
	
	
	\printAffiliationsAndNotice{\icmlEqualContribution} 
	
	\begin{abstract}
		Image processing is a fundamental task in computer vision, which aims at enhancing image quality and extracting essential features for subsequent vision applications. Traditionally, task-specific models are developed for individual tasks and designing such models requires distinct expertise. Building upon the success of large language models (LLMs) in natural language processing (NLP), there is a similar trend in computer vision, which focuses on developing large-scale models through pretraining and in-context learning. This paradigm shift reduces the reliance on task-specific models, yielding a powerful unified model to deal with various tasks. However, these advances have predominantly concentrated on high-level vision tasks, with less attention paid to low-level vision tasks. To address this issue, we propose a universal model for general image processing that covers image restoration, image enhancement, image feature extraction tasks, \textit{etc}. Our proposed framework, named PromptGIP, unifies these diverse image processing tasks within a universal framework. Inspired by NLP question answering (QA) techniques, we employ a visual prompting question answering paradigm. Specifically, we treat the input-output image pair as a structured question-answer sentence, thereby reprogramming the image processing task as a prompting QA problem. PromptGIP can undertake diverse \textbf{cross-domain} tasks using provided visual prompts, eliminating the need for task-specific finetuning. Capable of handling up to 15 different image processing tasks, PromptGIP represents a versatile and adaptive approach to general image processing. While PromptGIP has demonstrated a certain degree of out-of-domain task generalization, further research is expected to fully explore its more powerful capability.
	\end{abstract}
	
	\begin{figure}[htbp]
		\centering
		\includegraphics[width=0.99\linewidth]{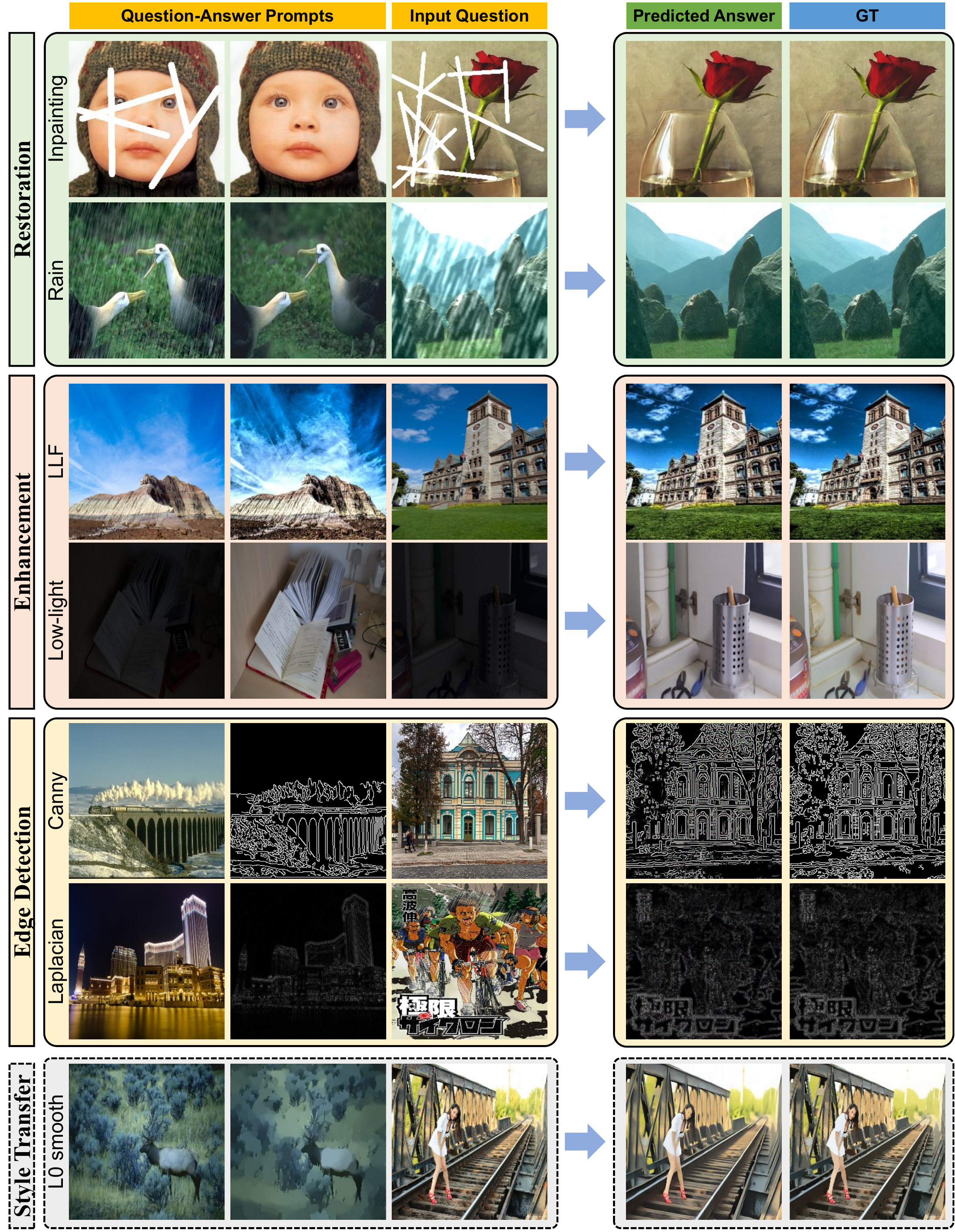}
		\vspace{-5pt}
		\caption{PromptGIP is a universal framework for general image processing. It can accomplish diverse tasks with distinct output domains, including image restoration, enhancement and edge detection. It has demonstrated a certain level of generalization for out-of-domain tasks (marked in dashed lines).}
		\vspace{-10pt}
	\end{figure}
	
	\vspace{-5pt}
	\section{Introduction}
	\label{sec:Introduction}
	\vspace{-2pt}
	
	Image processing encompasses a set of fundamental tasks that are aimed at direct manipulation and enhancement of image pixel-level information. These tasks are primarily focused on improving image quality and extracting basic image features, including but not limited to image restoration, image enhancement, image filtering, and image feature extraction. They provide a solid foundation for subsequent analysis, recognition, and comprehension of visual content within images. To address diverse image processing requirements, practitioners have traditionally resorted to developing specialized task-specific models. Consequently, achieving a particular objective often demands the utilization of different independent or combined models. 
	
	\begin{figure*}[htbp]
		\centering
		\includegraphics[width=0.8\linewidth]{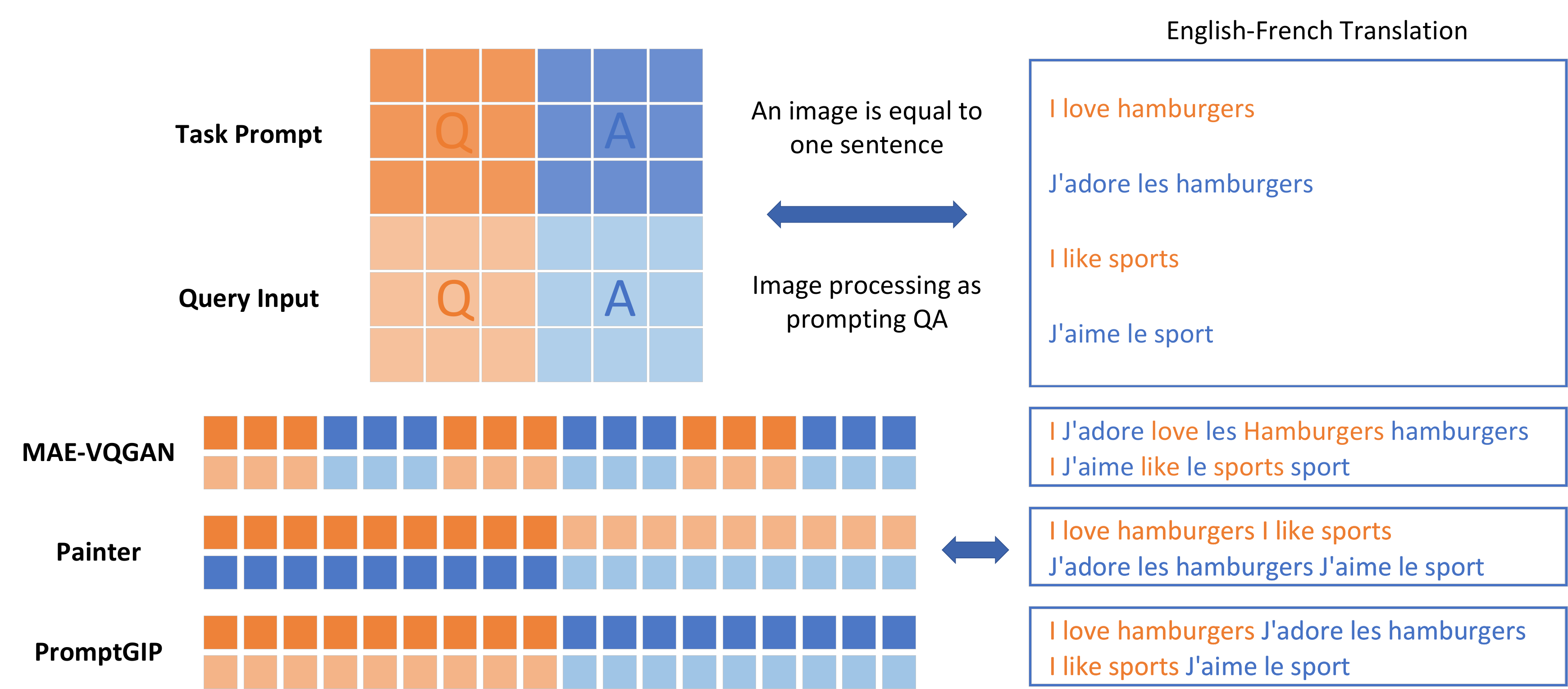} 
		\vspace{-5pt}
		\caption{Analogous to NLP tasks, various image processing tasks can be unified into a general visual prompting QA paradigm: given a pair of image prompt, the model can process the query image based on the prompts. MAE-VQGAN fragments image tokens and arrange them in an interleaved fashion. It disrupts the continuity and contextual understanding of the image content. Painter adopts a Q-Q-A-A organizational structure, which is not aligned with the QA paradigm. This misalignment can lead to inefficiencies in learning.}
		\label{fig:organization}
		\vspace{-10pt}
	\end{figure*}
	
	In recent years, a significant trend has emerged towards the development of general large-scale models. This paradigm shift involves extensive pretraining on massive datasets and interactive in-context learning techniques, leading to the creation of a unified, powerful model capable of handling multiple tasks. For example, large language models (LLMs), especially the GPT series models \cite{GPT2,GPT3}, have successfully unified most tasks in the natural language processing (NLP) field and achieved exceptional performance. Similar exploration has also been observed in the field of computer vision. Meta AI Research introduced a Segment Anything Model (SAM) \cite{SAM} for image segmentation. Through large-scale pretraining, SAM achieves remarkable zero-shot generalization performance in various scenarios. In other computer vision fields, a quantity of large foundation models have also been proposed, such as Inpainting Anything Model (IAM) \cite{IAM}, Track Anything Model (TAM) \cite{TAM}, InternImage \cite{InternImage}, and InternVideo \cite{InternVideo}. These advancements carry profound implications for the realization of artificial general intelligence (AGI).
	
	However, current focus of large models primarily lies in the domain of high-level vision. Low-level vision has received relatively little attention. While some newly-proposed methods, \textit{e.g.}, MAE-VQGAN \cite{MAEVQGAN} and Painter \cite{Painter}, have involved a few classic low-level vision tasks, their main focus remains on high-level vision tasks. Furthermore, these methods encounter challenges in dataset selection, model design, and training paradigms, making them unable to directly adapt to the low-level vision. 
	
	In this paper, we present a universal model for \textit{general image processing} by thoroughly examining the characteristics of low-level vision tasks and analyzing the limitations of existing in-context learning models in computer vision. Unlike prior literature that predominantly focused on image restoration tasks, our proposed model expands its scope to encompass image restoration, image enhancement, and image feature extraction. These tasks all belong to the domain of image processing, but their objectives and \textit{output domains} are distinct. Specifically, image restoration aims to recover the original clean and natural image from a degraded image, such as denoising and deblurring. Image enhancement focuses on improving the visual quality of the image by enhancing contrast, brightness, color tones, and textures. Image feature extraction, like edge detection, focuses on extracting the basic features from the image. Due to the different output representations, conventional image restoration models cannot accomplish these diverse \textbf{cross-domain} tasks by simply expanding the training data within a streamlined framework. To mitigate the ambiguity across different output domains, substantial task-specific retraining is needed. To address the diverse challenges of general image processing tasks, we adopt a visual prompting question answering paradigm, which utilizes paired visual prompts to precisely indicate the tasks to be accomplished. Our universal model, namely PromptGIP, can effectively handle \textbf{up to 15} various image processing tasks, providing a more versatile solution for low-level vision. The experiments also indicate that in-context learning enables the model to exhibit preliminary generalization for out-of-domain tasks.
	
	\vspace{-5pt}
	\section{Related Work}
	\vspace{-2pt}
	
	\noindent\textbf{Image Restoration and Beyond.} Over the past decade, single-purpose image restoration methods, dedicated to recover the original clean and natural image from degraded observation, have garnered substantial research attention. Numerous representative approaches have found applications across various domains, including denoising \cite{DNCNN}, deblurring \cite{DeblurGAN}, and deraining \cite{MPRNet}, among others.
	However, the inherent limitation of these techniques lies in their reliance on specialized datasets and the tailored network architectures. Consequently, their generalization ability remains unsatisfactory, falling notably short of generality.
	Moreover, image enhancement algorithms, like low-light enhancement \cite{LOL}, present significant demands and applications. Paradoxically, most researchers tend to concentrate solely on a specific augmentation methodology, such as simply enlarging the training data, to seek for more robust generalization. Differently, we advocate for a paradigm shift, expanding the purview beyond image restoration to embrace image enhancement and other image processing tasks. Besides, we propose a unified framework capable of collectively tackling all these tasks. This pioneering approach markedly enhances the universality of low-level vision foundation models, bridging the gap between disparate domains.
	
	\noindent\textbf{Visual In-Context Learning.} In NLP, the GPT (Generative Pretrained Transformer) series models, such as GPT-2 and GPT-3 \cite{GPT2,GPT3}, have achieved significant success in unifying various NLP tasks. By providing a prompt or designing an in-context example, which is usually a task-specific instruction or question, GPT can be transformed into a task-specific question-answering model without the need for extensive retraining or fine-tuning. 
	In vision, a few works -- MAE-VQGAN \cite{MAEVQGAN} and Painter \cite{Painter}, have begun harnessing the flexibility afforded by in-context learning to unify diverse vision tasks. By constructing grid-like prompts, they exhibit commendable performance on high-level tasks like semantic segmentation. However, their efficacy has been less pronounced in low-level domains, failing to exploit the full potential of in-context learning.
	We claim that this discrepancy may be attributed to the distinct nature of low-level vision tasks, which involve pixel-wise image manipulation, in contrast to the high-level tasks that demand comprehension across varying levels of abstraction.

	\noindent\textbf{Multi-task Learning for Image Processing.} 
	Multi-Task Learning (MTL) aims to train a single model to concurrently handle multiple image processing tasks.
	Traditionally, MTL approaches have predominantly focused on image restoration, and they can be broadly categorized into two streams. 
	BSRGAN \cite{BSRGAN} and RealESRGAN \cite{RealESRGAN} adopt a data-centric approach. They propose to employ models with significant parameter complexity and utilize complicated degradation models to generate ample training data. DASR \cite{DASR} and AirNet \cite{AirNet}, on the other hand, adopt a model-centric approach. They design specialized modules to implicitly capture diverse degradations and exploit them as conditions for achieving MTL.
	Beyond these approaches, ProRes \cite{ProRes} and PromptIR \cite{PromptIR} leverage prompts as a form of guidance or condition, enabling MTL for three (denoising, rain removal, and fog removal) tasks, or five (denoising, deraining, deblurring, low-light enhancement, and defogging) tasks.
	Despite these contributions, existing methodologies remain limited in their ability to tackle a modest number of MTL tasks, typically up to five. In contrast, our proposed approach breaks this ceiling by achieving MTL across more than ten distinct tasks (denoising, deblurring, deJPEG, dering, deraining, defogging, deraining, inpainting, low-light enhancement, local Laplacian filtering, and edge detection).
	
	\begin{figure*}[htbp]
		\centering
		\includegraphics[width=0.8\linewidth]{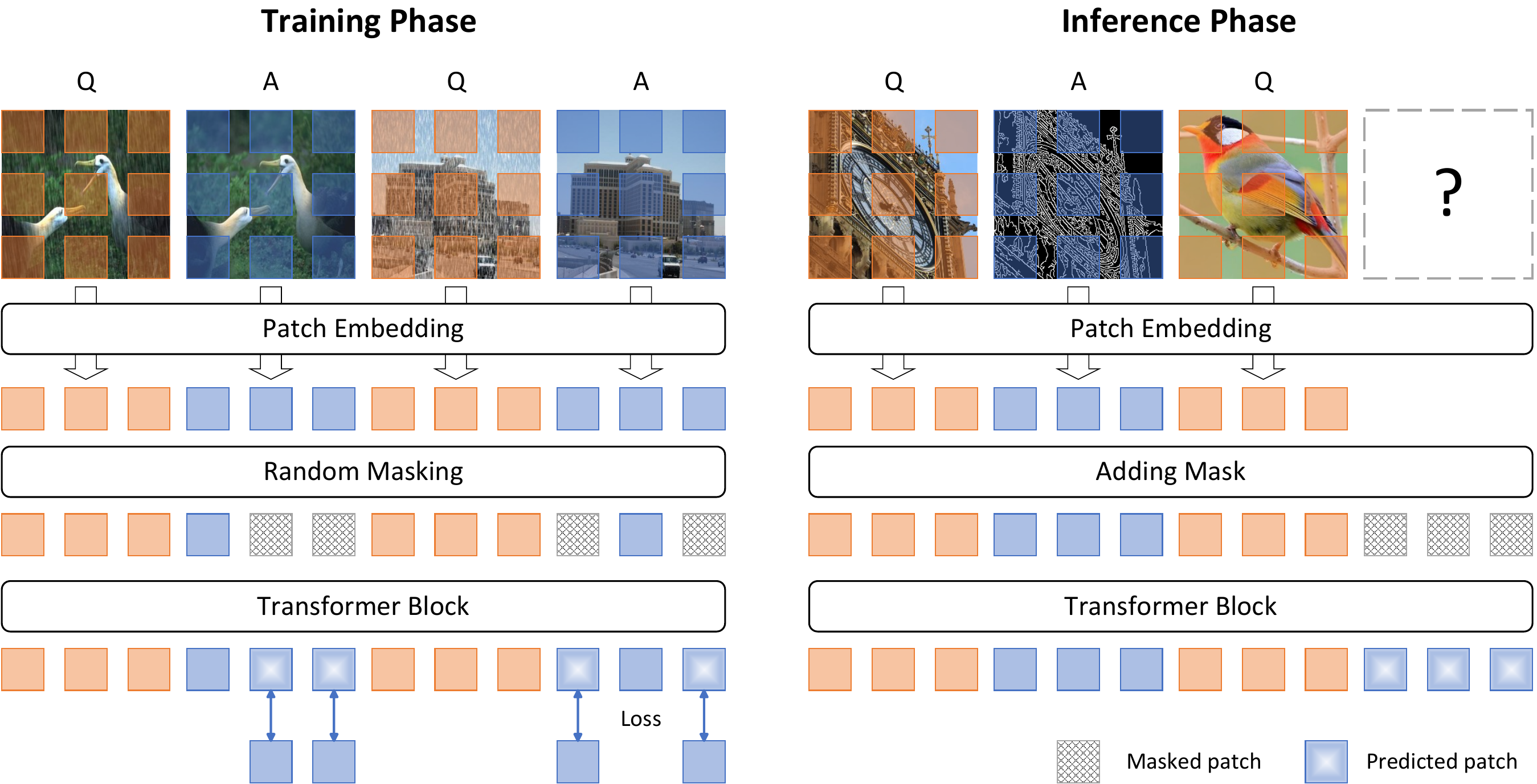}
		\caption{We structure the input and output images as a ``Q-A-Q-A'' sequence. During training, the answer images (A) are randonly masked and predicted. For inference, PromptGIP can execute proper processing to the question image according to the prompt pairs.}
		\label{fig:framework}
		\vspace{-10pt}
	\end{figure*}
	
	\vspace{-5pt}
	\section{Method}
	\vspace{-2pt}
	\subsection{Image Processing as Visual Question Answering}
	Compared to high-level vision tasks, low-level vision tasks necessitate meticulous pixel-level adjustments, demanding architectures that excel in processing intricate details. These tasks encounter diverse input/output domains, characterized by various degradations and complex operations. These challenges underscore the complexity and non-trivial nature of general image processing.
	
	Inspired by the success of prompting in NLP \cite{liu2023pre}, we propose to unify the general image processing problem as the visual prompting question answering (QA) paradigm, as illustrated in Fig. \ref{fig:organization}. In QA, the objective is to process a given context, such as a paragraph or document, and accurately generate the correct answer in response to specific questions related to that context. Building upon this concept, we adapt the QA paradigm to image processing.
	
	In our design, we view an image as a ``question'' (Q) or an ``answer'' (A). When inference, the model $G$ is initially provided with input-output image pairs ($P_Q$ and $P_A$), which serve as essential task prompts, much like the given context in QA tasks. These image pairs play a pivotal role in guiding the model's image processing operations. To process a new targeted input image $X_Q$, we encode it as the query ``question'' to be answered.
	The provided input-output image pairs then serve as contextual prompts, enabling the model to gain insightful cues to generate the desired output. With this knowledge, the model executes the appropriate image processing operations to produce the predicted ``answer'' $Y_A$:
	\begin{equation}
	Y_A = G(X_Q | \{P_Q, P_A\}).
	\end{equation}
	
	An illustrative example is shown in Fig. \ref{fig:framework}. The content of the prompts for the model is represented in the form of ``question''-``answer'' image pairs. For instance, when the input prompt is a ``rainy''-``rain-free'' image pair, the model will perform rain removal on the target input image. If the answer in the prompt is related to image edges, the model will conduct edge detection operations on the query image, producing the corresponding edge image as the output.
	
	Notably, PromptGIP is capable of handling tasks with distinct output domains, which was not achievable with previous image restoration methods. The output domain of image restoration is the natural image space; image enhancement involves transformations in image brightness, color tones, or styles; while image edge detection outputs edge features, not the RGB image space. Our approach unifies these different tasks within a unified framework.
	
	\vspace{-5pt}
	\subsection{Masked Visual Prompting Paradigm}
	\vspace{-2pt}
	Masked image modeling has emerged as a promising self-supervised technique for learning valuable visual representations. Following \cite{MAE}, we implement a similar masked autoencoding approach in our training process. As depicted in Fig. \ref{fig:framework}, we initially structure the input and output images within a ``Q-A-Q-A'' sequence. Then, we introduce random masking to certain portions of the answer images, prompting the model to reconstruct these masked patches from the unmasked counterparts. This procedure employs a mask ratio of 85\%. It is pivotal to note that our organizational framework distinguishes itself from prior works \cite{MAEVQGAN,Painter} in its more rational and effective design. More analyses are described in Sec. \ref{sec:discussion}.
	
	During the training phase, our approach leverages a diverse dataset comprising input-output image pairs, where each pair corresponds to a distinct image processing goal, including restoration, enhancement, and edge detection. Notably, each primary task encompasses various sub-tasks that further enrich the model's understanding. Throughout this process, the model is trained to grasp the intrinsic correlations between the Q-A image pairs. During the inference stage, we assemble an input-output pair as a task prompt, guiding the model to execute tailored operations. By providing an input question image alongside a fully masked image, the model generates the intended answer image in correspondence with the question image.

	\subsection{Further Discussion}\label{sec:discussion}
	\noindent\textbf{Comparison with image restoration models.}
	Earlier research primarily focused on crafting specialized models tailored to specific tasks, such as SRCNN \cite{SRCNN} for super-resolution, DnCNN \cite{DNCNN} for denoising, and Deblur-GAN \cite{DeblurGAN} for deblurring. While effective within constrained scenarios, these task-specific models possess limited generalization capability. Recent attention has pivoted toward all-in-one restoration methods \cite{AirNet,RealESRGAN}. These approaches leverage multi-task learning techniques to construct models that are able to handle diverse restoration tasks, thereby circumventing the need for task-specific finetuning. Nonetheless, these models are often limited within predefined application domains. They fall short in producing alternative representations like stylistic images or image edges. Several concurrent works \cite{ProRes,PromptIR} have embraced the concept of prompt learning, but still concentrate on image restoration tasks. They propose to incorporate learnable prompts as degradation embeddings to guide the restoration process. However, it is worth noting that general image processing encompasses more than just restoration tasks. In this context, PromptGIP demonstrates a remarkable adaptability across a wide spectrum of low-level vision tasks, liberating it from the constraints of a singular output domain.
	
	\begin{figure}[htbp]
		\centering
		\includegraphics[width=0.99\linewidth]{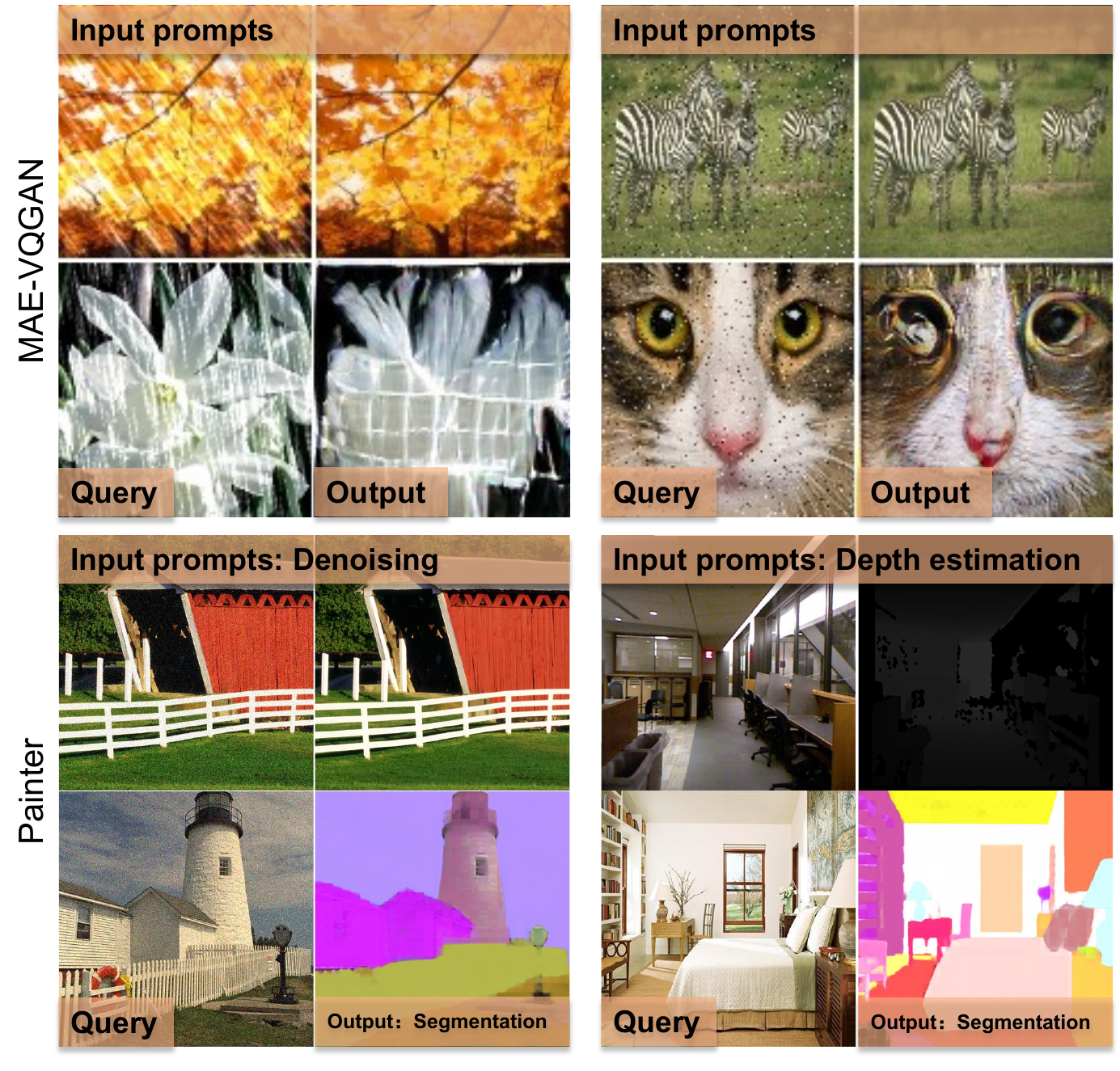} 
		\vspace{-10pt}
		\caption{The drawbacks of existing methods. MAE-VQGAN fails to produce high-quality images. The prompts of Painter do not actually work well.}
		\label{fig:MAEVQGAN_fail}
		\vspace{-15pt}
	\end{figure}
	
	\noindent\textbf{Comparison with existing visual prompting models.}
	Two novel visual prompting techniques, MAE-VQGAN~\cite{MAEVQGAN} and Painter~\cite{Painter}, have emerged for addressing various tasks. MAE-VQGAN employs a masked autoencoder for pretraining. Unlike predicting masked pixels, it predicts visual tokens from a pretrained VQGAN codebook. The training process involves the ImageNet dataset and the collected CVF dataset, comprising a diverse array of figures from computer vision papers. Painter combines pairs of images to predict the output domain through the masked image modeling. It encompasses high-level tasks and a few low-level tasks.
	
	\textit{Differences with MAE-VQGAN.} MAE-VQGAN diverges from the question-answering paradigm. As illustrated in Fig.~\ref{fig:organization}, MAE-VQGAN utilizes crude images extracted from the ImageNet/CVF dataset during MAE training, and stitches paired images as a whole image for inference. This straightforward and coarse data organization scheme deviates from the QA framework. Specifically, during the training phase, the model lacks the capability to differentiate whether a given visual token corresponds to a ``Question'' or an ``Answer'', leading to an interleaved and ambiguous input-output encoding. In contrast, our approach is firmly grounded in an explicit QA paradigm, enabling precise pixel-level prediction. In addition, MAE-VQGAN choose to predict VQGAN tokens rather than pixels, which results in subpar fidelity of reconstructed content, as exemplified in Fig.~\ref{fig:MAEVQGAN_fail}. On the contrary, our framework excels in pixel-wise prediction with visually compelling outcomes.

	\textit{Differences with Painter.} Painter predominantly targets high-level vision tasks, encompassing segmentation, depth estimation, and keypoint detection, while it also addresses limited low-level task. Painter only draws training data from just seven specific datasets. This potentially induces a propensity for excessive alignment with these datasets, which can easily result in a concomitant risk of overfitting. Furthermore, the implementation of Painter employs a ``Q-Q-A-A'' sequence for encoding prompt and query images (see Fig. \ref{fig:organization}). Such a design yields unexpected behaviors in Painter's response to task prompts. Extensive tests on Painter revealed disparities in its prompt mechanism compared to anticipated outcomes. As in Fig. \ref{fig:MAEVQGAN_fail}, when provided prompts linked to denoising and depth estimation, it unexpectedly executes segmentation tasks. This phenomenon hints at the model's inclination to memorize specific datasets rather than effectively leveraging the provided prompts. We conjecture that this problem might stem from the limited range of training tasks.
	
	\begin{figure*}[htbp]
		\centering
		\includegraphics[width=0.8\linewidth]{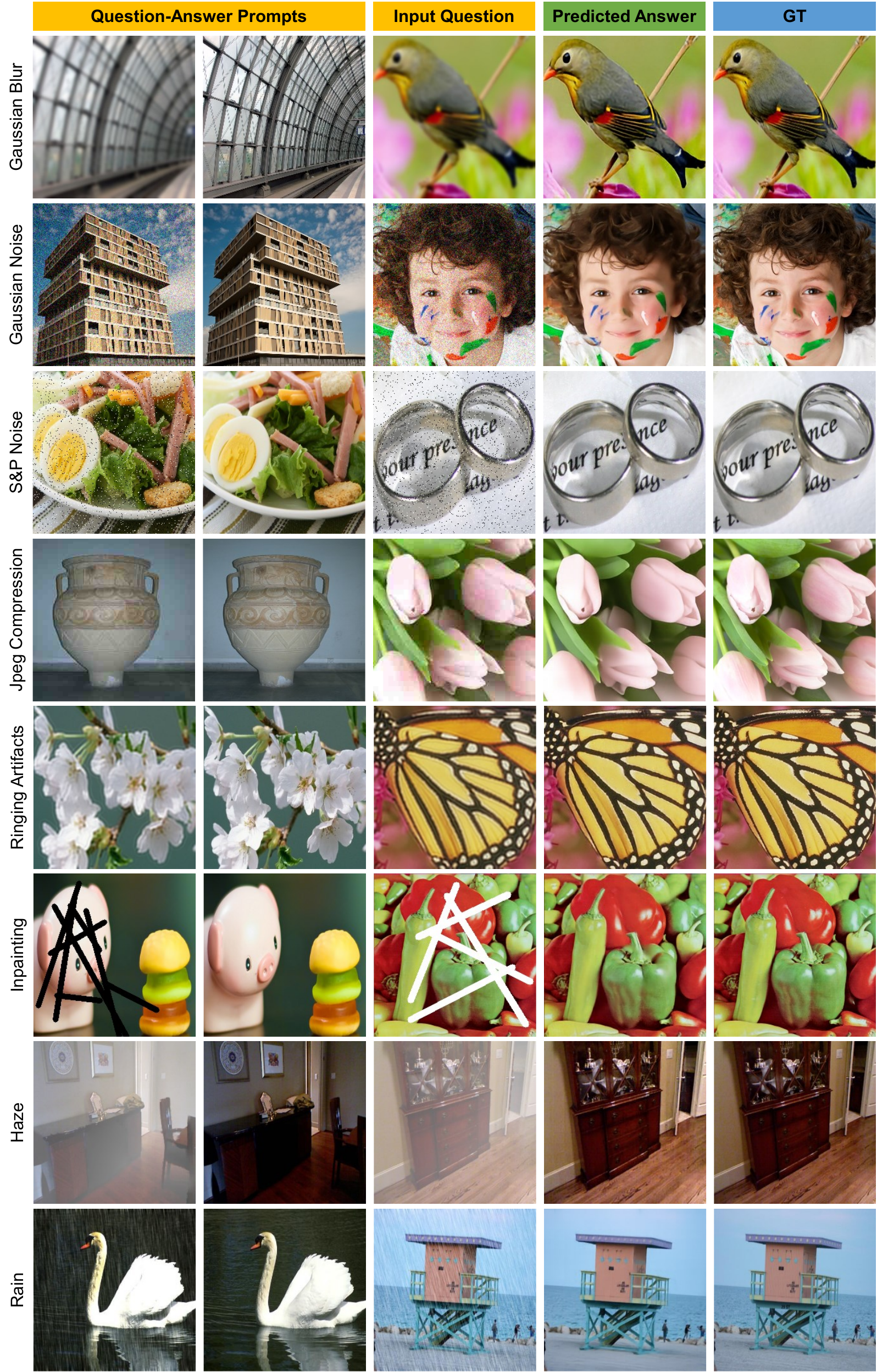} 
		\caption{Visual results of PromptGIP on all-in-one multi-task restoration.}
		\label{fig:visual_result1}
	\end{figure*}
	
	\begin{table*}[htbp]
		\centering
		\renewcommand{\arraystretch}{1.2}
		\caption{Quantitative results (PSNR/SSIM) on image restoration tasks. $\star$: trained with only restoration tasks. $\spadesuit$: trained with all image processing tasks. $\dagger$: public released model.}
		\resizebox{\textwidth}{!}{%
			\begin{tabular}{r|ccccccccccc}
				\hline
				&
				Gaussian Noise &
				Poisson Noise &
				S\&P Noise &
				Gaussian Blur &
				JPEG &
				Ringing &
				R-L &
				Inpainting &
				Simple Rain &
				Complex Rain &
				Haze \\ \hline
				Real-ESRGAN\textsuperscript{$\dagger$} &
				25.38/0.7997 &
				26.57/0.8472 &
				21.50/0.5884 &
				21.49/0.6263 &
				25.21/0.8058 &
				24.64/0.7834 &
				21.71/0.6548 &
				14.06/0.7084 &
				16.10/0.5989 &
				21.01/0.6705 &
				11.86/0.6346 \\ 
				Restormer\textsuperscript{$\star$} &
				28.66/0.8731 &
				31.31/0.9317 &
				36.12/0.9851 &
				24.24/0.7537 &
				26.65/0.8391 &
				27.14/0.8561 &
				30.53/0.9306 &
				27.77/0.9289 &
				29.68/0.9476 &
				24.26/0.8369 &
				14.83/0.7382 \\
				ViT-large\textsuperscript{$\star$} &
				24.67/0.7804 &
				25.39/0.8152 &
				23.71/0.7335 &
				22.17/0.6413 &
				24.76/0.7920 &
				23.89/0.7463 &
				24.09/0.7335 &
				23.11/0.7662 &
				23.21/0.7620 &
				23.04/0.7788 &
				24.91/0.8565 \\ \hline
				Restormer\textsuperscript{$\spadesuit$} &
				25.27/0.7634 &
				27.22/0.8535 &
				27.84/0.8811 &
				21.71/0.6078 &
				23.90/0.7606 &
				23.61/0.7261 &
				23.18/0.7120 &
				24.19/0.8615 &
				22.68/0.7879 &
				20.39/0.6930 &
				7.22/0.1395 \\ 
				Painter\textsuperscript{$\spadesuit$} &
				24.17/0.7468 &
				24.63/0.7792 &
				24.75/0.7903 &
				22.36/0.6477 &
				23.97/0.7458 &
				24.21/0.7531 &
				24.56/0.7728 &
				22.95/0.7455 &
				23.35/0.7493 &
				22.81/0.7710 &
				20.60/0.8250 \\ 
				PromptGIP\textsuperscript{$\spadesuit$} &
				26.22/0.8167 &
				27.29/0.8590 &
				27.49/0.8804 &
				22.77/0.6911 &
				25.38/0.7978 &
				25.45/0.8079 &
				26.79/0.8506 &
				25.02/0.8401 &
				25.46/0.8399 &
				24.08/0.8322 &
				24.32/0.9020 \\ \hline
			\end{tabular}
		}
		\vspace{-15pt}
		
		\label{tab:restoration}
	\end{table*}
	
	\vspace{-5pt}
	\section{Experiments and Analysis}
	\vspace{-2pt}
	\subsection{Image Processing Task Settings}
	To show the versatility of our proposed method, we incorporate up to \textbf{15} tasks including diverse image restoration, image enhancement, and image edge detection tasks into our experiments. These tasks have their distinct output domains.
	
	\noindent\textbf{Image restoration.} We consider 10 degradation types: Gaussian noise, Gaussian blur, Poisson noise, salt \& pepper noise, jpeg compression, ringing artifacts, R-L algorithm \cite{richardson1972bayesian}, inpainting, haze, and rain. For the first eight types, we directly introduce corresponding distortions to the ImageNet \cite{ImageNet} dataset to create degraded-clean pairs. We collect a composed dataset (Common528) for testing, which consists of commonly-used datasets: Set5 \cite{Set5}, Set14 \cite{Set14}, BSDS100 \cite{BSDS100}, Manga109 \cite{Manga109}, Urban100 \cite{Urban100}, General100 \cite{FSRCNN}, and DIV2K-Valid \cite{DIV2K}. For dehazing, we utilize the ITS training set of RESIDE dataset \cite{RESIDE}. For rain removal, we employ two types of rain addition models: Simple Rain Model and Complex Rain Model. The former is a straightforward additive rain model, directly synthesized on the ImageNet dataset; while the latter utilizes Rain13K \cite{MPRNet}, including an assortment of diverse rain models.
	
	\noindent\textbf{Image enhancement.} We employ two enhancement tasks: low-light image enhancement (LLE) and local Laplacian filtering (LLF). For LLE, the LOL dataset \cite{LOL} is adopted for training. For LLF, we apply local Laplacian filter \cite{LLF} on the expert-C retouched images of Adobe-MIT Fivek dataset \cite{FiveK}, forming the requisite input-output pairs. LLF is a multi-scale operator for edge-preserving detail enhancement.
	
	\noindent\textbf{Image edge detection.} Two acknowledged image edge detection operators, the Canny and Laplacian operators, are investigated. The ImageNet dataset forms the basis for creating input-output training pairs.

	All these 15 diverse tasks are amalgamated within a unified setting. PromptGIP excels in accommodating these tasks under a cohesive framework with one single training phase.

	\setlength\tabcolsep{5pt}
	\begin{table}[t]
		\centering
		\caption{Quantitative results on image enhancement and image edge detection. $\star$: single models trained with individual tasks. $\spadesuit$: trained with all image processing tasks.}
		\scriptsize
		\begin{tabular}{c|cccc|cc}
			\hline
			& \multicolumn{2}{c}{LLE (LOL dataset)}       & \multicolumn{2}{c|}{LLF} & Canny   & Laplacian \\
			& PSNR$\uparrow$ & SSIM$\uparrow$ & PSNR$\uparrow$       & SSIM$\uparrow$        & MAE$\downarrow$     & MAE$\downarrow$       \\ \hline
			ViT-large\textsuperscript{$\star$} & 13.37         & 0.4892        & 25.42      & 0.8948      & 36.5290 & 1.4655    \\ \hline
			Painter\textsuperscript{$\spadesuit$} & 19.47         & 0.7491        & 23.87      & 0.8451      & 33.7188 & 5.4518    \\ 
			PromptGIP\textsuperscript{$\spadesuit$} & 20.30         & 0.8026        & 26.11      & 0.9107      & 21.4376 & 3.7852    \\ \hline
		\end{tabular}
		\vspace{-15pt}
		
		\label{tab:enhancement&edge}
	\end{table}
	
	\vspace{-5pt}
	\subsection{Implementation Details}
	\vspace{-2pt}
	A vanilla vision Transformer (ViT-large) \cite{ViT} is adopted as the backbone architecture. During training, the model processes sequences of four 256 $\times$ 256 images in a ``Q-A-Q-A'' pattern, resulting in a 4 $\times$ 256 $\times$ 256 total input resolution. $L_1$ loss is utilized as the loss function. For optimization, AdamW \cite{AdamW} optimizer with a cosine learning rate scheduler is employed. The base learning rate is $1e-4$. The batch size is 48. We use 8 Tesla V100 GPUs for training. A total of 50 epochs are executed. For testing Painter and PromptGIP, we construct 20 image prompts for each task and report the best results.

	\vspace{-5pt}
	\subsection{Experiments}
	\vspace{-2pt}
	Currently, there is no existing unified network that can comprehensively address all the aforementioned tasks in an all-in-one manner. For instance, previous image restoration models are incapable of handling image edge detection task. For reference, we train a ViT-large model and a Restormer model \cite{Restormer} using the same training policy on multiple restoration tasks. We retrain the Painter \cite{Painter} model with all tasks as PromptGIP. We also report the results of Real-ESRGAN \cite{RealESRGAN}, which is proposed to handle various complex restoration. Due to differences in the performance of various architectures, absolute numerical comparisons would be unfair. We have opted for the simplest ViT structure, thus it is more fair to focus on a direct comparison with the ViT and Painter. 
	
	Moreover, achieving state-of-the-art performance on every task is not the purpose of this paper. Our primary focus revolves around examining the effects and capability of prompt learning in the context of general image processing. We can focus more on functional outcomes rather than numerical results. The metrics are evaluated on RGB channels. 
	
	\begin{figure}[t]
		\centering
		\includegraphics[width=0.99\linewidth]{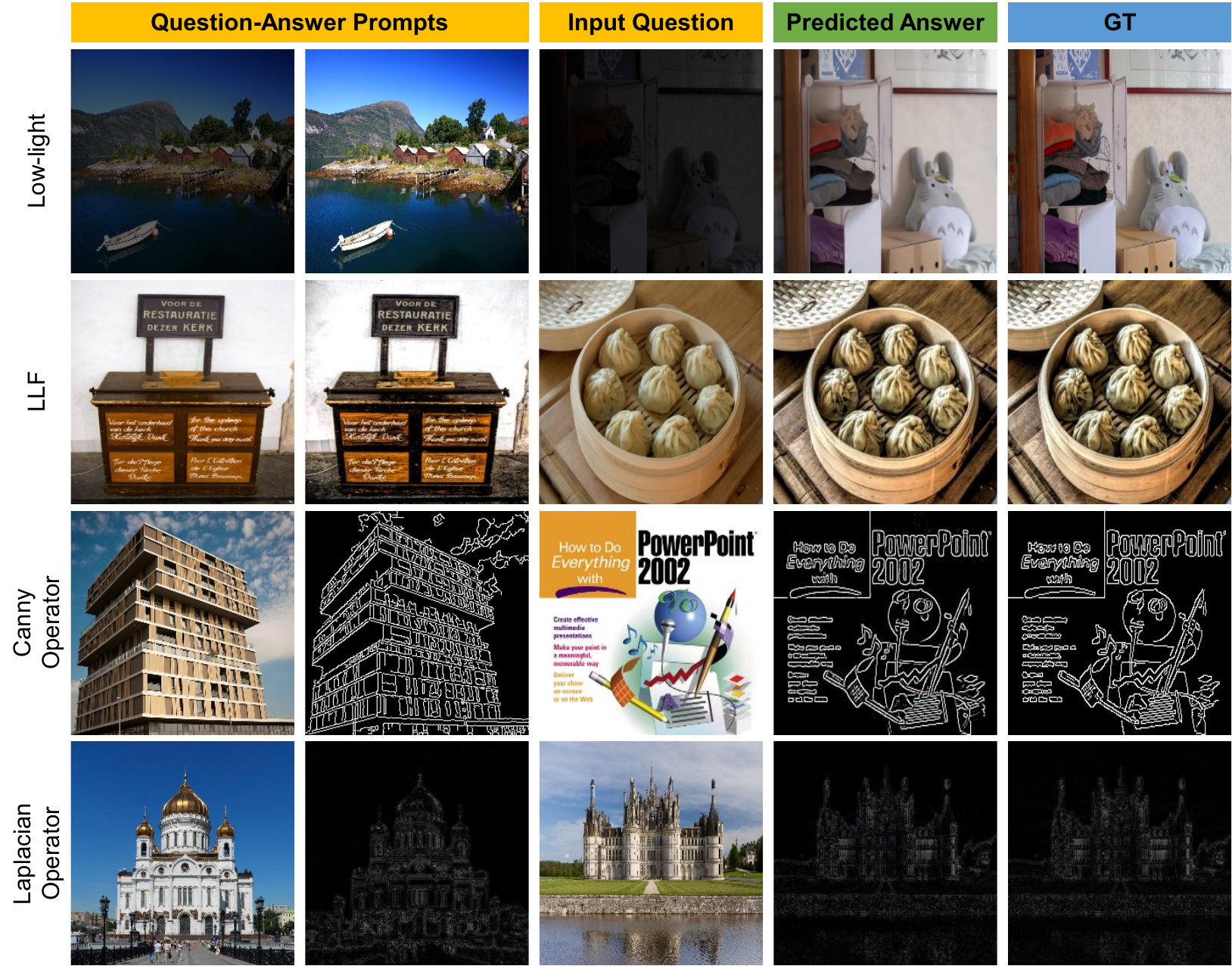} 
		\vspace{-10pt}
		\caption{Visual results of PromptGIP on image enhancement and edge detection tasks.}
		\label{fig:visual_result2}
		\vspace{-15pt}
	\end{figure}

\begin{figure}[htbp]
		\centering
		\includegraphics[width=0.99\linewidth]{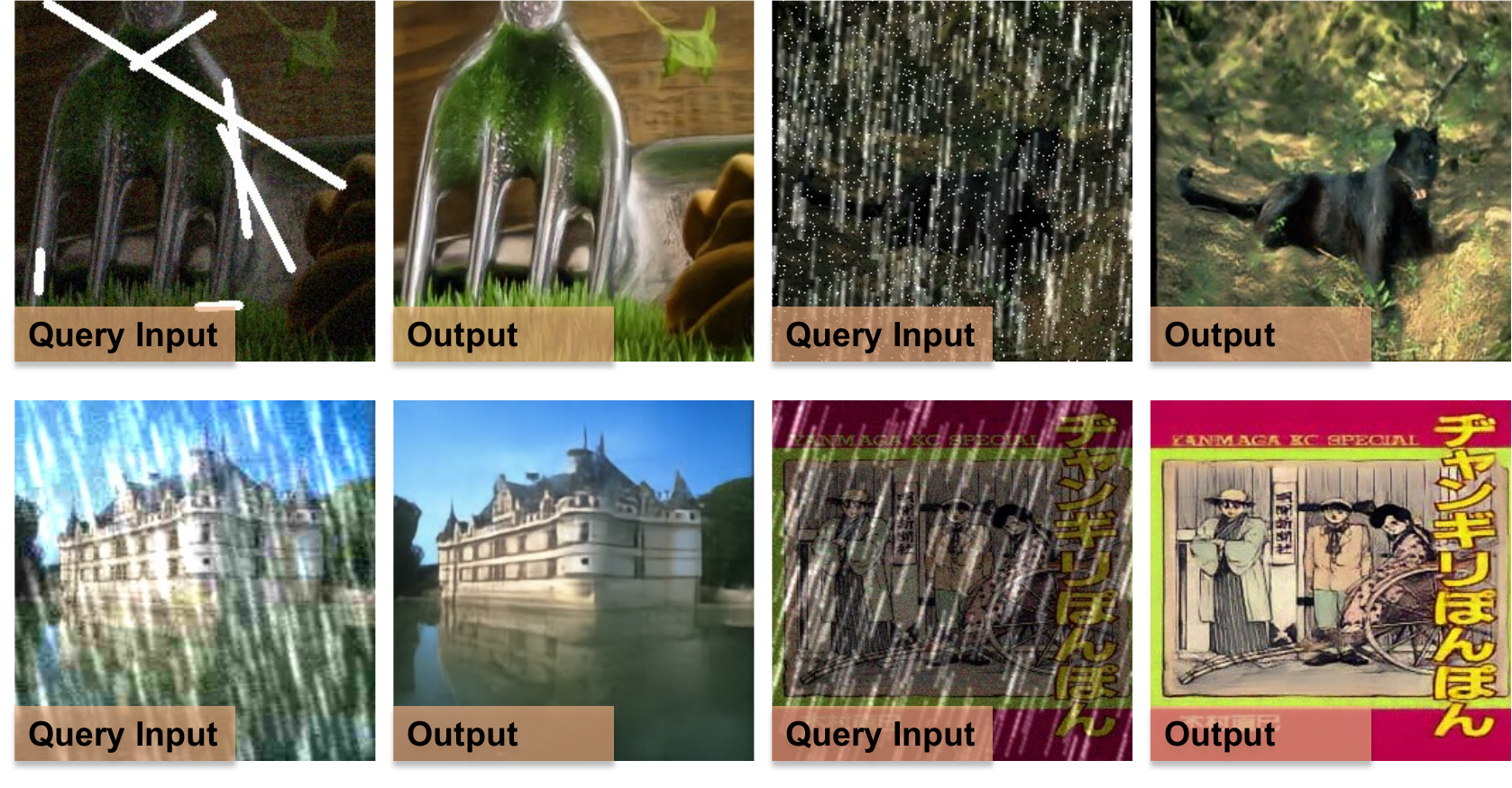}
		\vspace{-10pt}
		\caption{Results of mixed degraded images.}
		\label{fig:mix_deg}
		\vspace{-12pt}
	\end{figure}

	\noindent\textbf{Results.}
	Illustrated in Fig. \ref{fig:visual_result1} and \ref{fig:visual_result2}, PromptGIP proficiently addresses a range of image processing tasks using different input prompts. These tasks encompass multiple-degradation restoration, enhancement, and edge detection. These tasks entail distinct output representations, a level of complexity that lies beyond the capability of existing image restoration methods. PromptGIP yields impressive visual results in diverse tasks. In the training process, we introduced mixed degradation scenarios to further challenge the model's restoration capability, with results presented in Fig. \ref{fig:mix_deg}.
	
	Quantitative results for restoration are detailed in Tab. \ref{tab:restoration}, where PromptGIP demonstrates appealing performance across 10 restoration tasks using a vanilla ViT backbone. Compared to the original ViT model and Painter, prompt learning demonstrates a significant enhancement in model performance, resulting in improved restoration and multitasking capability. PromptGIP also surpasses the performance of Real-ESRGAN, a model specifically crafted for blind image restoration. PromptGIP achieves higher quantitative score than Restormer on complex derain and dehaze tasks. Restormer achieves superior quantitative scores on other degradations. This can be attributed to Restormer's advanced architecture tailored for restoration tasks. 
	
	PromptGIP also succeeds in enhancing low-light images and emulating image operators. Notably, earlier methods struggle to simultaneously realize all these tasks within a single framework, due to variances in output domain representations. However, PromptGIP, when provided with proper prompts, effectively executes a wide spectrum of image processing tasks within a singular, streamlined network, as depicted in Fig. \ref{fig:visual_result2}. For image edge detection, the Canny operator produces clear and well-defined edges, while Laplacian operator tends to produce thicker and noisier edges. Despite these intricacies, PromptGIP successfully discerns and faithfully simulates the distinct behaviors of both operators, underscoring its impressive adaptability. The numerical results are shown in Tab. \ref{tab:enhancement&edge}.
	
	\noindent\textbf{Effectiveness of the QA paradigm and masked training.}
	We further validate the efficacy of our newly proposed QA paradigm and the masked training strategy. Unlike the encoding order of Q-Q-A-A used in Painter, our PromptGIP employs a Q-A-Q-A approach. Q-Q-A-A could dilute the model's focus and impair its ability to directly map questions to their relevant answers. Our paradigm significantly improves performance, as in Tab. \ref{tab:restoration} and \ref{tab:enhancement&edge}, which demonstrate superior outcomes in both image restoration and enhancement tasks. Additionally, we emphasize the necessity of our masked training strategy. During the training phase, PromptGIP randomly masks patches in the two ``answer'' images, in contrast to direct predicting where only the last ``answer'' image is masked. This methodology, as shown in Tab. \ref{tab:mask_strategy}, proves more effective across all tasks, particularly in image dehazing, where direct predicting struggles to yield satisfactory results. This outcome suggests that masked training not only enhances the model's capability in handling diverse tasks but also contributes to its generalization and stability.
	
	\begin{table}[t]
		\centering
		\caption{Effectiveness of the proposed QA paradigm and masked training strategy.}
		\resizebox{\linewidth}{!}{%
			\begin{tabular}{c|cc|cccc}
				\hline
				&
				Encoding Paradigm &
				Mask Strategy &
				\begin{tabular}[c]{@{}c@{}}Poisson Noise\\ PSNR$\uparrow$\end{tabular} &
				\begin{tabular}[c]{@{}c@{}}Haze\\ PSNR$\uparrow$\end{tabular} &
				\begin{tabular}[c]{@{}c@{}}LLF\\ PSNR$\uparrow$\end{tabular} &
				\begin{tabular}[c]{@{}c@{}}Laplacian\\ MAE$\downarrow$\end{tabular} \\ \hline
				Painter           & $\text{Q}_1$-$\text{Q}_2$-$\text{A}_1$-$\text{A}_2$ & $\text{A}_1 \& \text{A}_2$ & 24.63 & 20.60 & 23.87 & 5.4518  \\
				Direct predicting & $\text{Q}_1$-$\text{A}_1$-$\text{Q}_2$-$\text{A}_2$ & only $\text{A}_2$          & 26.30 & 18.57 & 25.38 & 11.5553 \\
				PromptGIP         & $\text{Q}_1$-$\text{A}_1$-$\text{Q}_2$-$\text{A}_2$ & $\text{A}_1 \& \text{A}_2$ & 27.29 & 24.32 & 26.11 & 3.7852  \\ \hline
			\end{tabular}
		}
		
		\label{tab:mask_strategy}
		\vspace{-15pt}
	\end{table}
	
	\begin{figure}[htbp]
		\centering
		\includegraphics[width=0.99\linewidth]{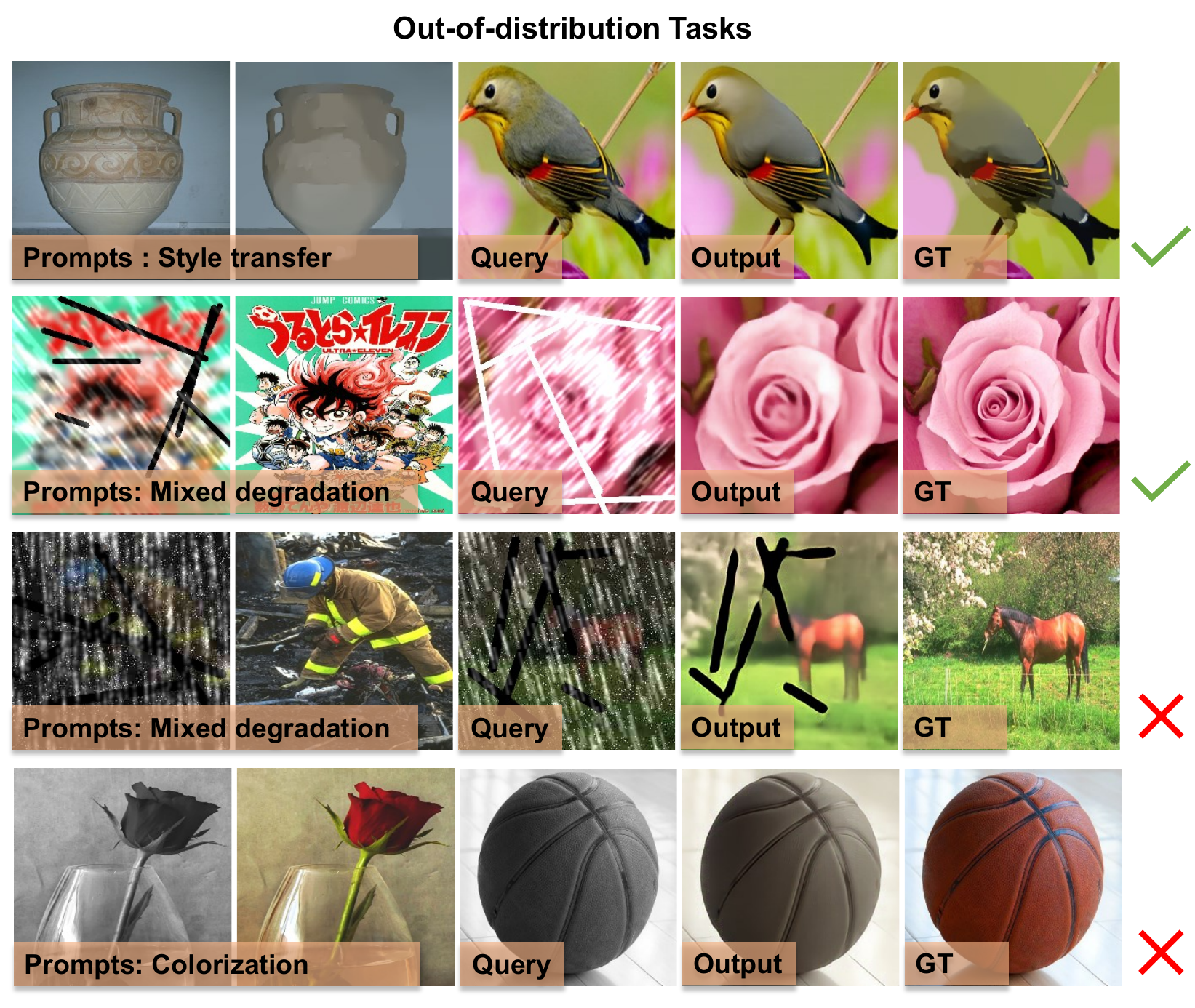}
		\vspace{-10pt}
		\caption{Although PromptGIP cannot perfectly deal with every out-of-distribution tasks, it has demonstrated a certain level of generalization capability.}
		\label{fig:outdistribution}
		\vspace{-15pt}
	\end{figure}
	
	\noindent\textbf{Exploration on out-of-distribution tasks.}
	To evaluate the model's capacity for generalization, we incorporate a set of diverse out-of-distribution tasks that are intentionally not encountered during the training phase, including mixed degradation restoration, colorization, and style transfer. The results are presented in Fig. \ref{fig:outdistribution}. We employ $L_{0}$ smooth filtering \cite{xu2011image} to conduct style transfer experiment. As shown in Fig. \ref{fig:outdistribution}, the model seems to understand the input prompt pair, yielding images with a discernible $L_{0}$ smooth filter style. While it occasionally succeeds in producing visually appealing reconstructed images, it encounters difficulties in effectively restoring unfamiliar mix degraded images when compared to seen degraded data. Additionally, we provide a grayscale-colorful image pair as a prompt, with the expectation that the model would apply colorization to the grayscale input. However, the model regrettably does not exhibit colorization behavior in response to this prompt. In summary, these observations highlight the model's capacity to discern the intended task from the prompt and endeavor to fulfill it, showcasing a certain level of generalization. It is essential to emphasize that the model's present capability does not extend to generating great ``emergent'' outcomes. These conclusions are in accordance with prior studies \cite{min2022rethinking,wei2023larger}.
	
	\vspace{-5pt}
	\section{Conclusion}
	\vspace{-2pt}
	We present PromptGIP, a versatile model designed to address a wide spectrum of image processing tasks. By adopting a unique visual prompting question answering paradigm, PromptGIP adeptly handles tasks like image restoration, enhancement, and edge detection. Our comprehensive experimental evaluations affirm PromptGIP's commendable capacity to interpret the implicit task cues embedded within visual prompts, yielding pertinent outputs that underscore a noteworthy level of generalization. This study sheds light on exploration and refinement of universal image processing models in the future.
	
	\vspace{-5pt}
	\section{Broader Impact}
	\vspace{-2pt}
	The proposed PromptGIP framework represents a new advancement in the field of image processing, offering a versatile and efficient solution for a wide range of low-level vision tasks. By adapting the training data, PromptGIP could facilitate new applications in medical imaging, environmental monitoring, and content creation, making high-quality image processing more accessible to non-experts and contributing positively to fields where image quality is crucial. Additionally, this research could inspire further interdisciplinary work, integrating concepts from natural language processing and computer vision to solve complex problems in novel ways.
	
	\bibliography{icml2024_bib}

\begin{thebibliography}{48}
\providecommand{\natexlab}[1]{#1}
\providecommand{\url}[1]{\texttt{#1}}
\expandafter\ifx\csname urlstyle\endcsname\relax
  \providecommand{\doi}[1]{doi: #1}\else
  \providecommand{\doi}{doi: \begingroup \urlstyle{rm}\Url}\fi

\bibitem[Agustsson \& Timofte(2017)Agustsson and Timofte]{DIV2K}
Agustsson, E. and Timofte, R.
\newblock Ntire 2017 challenge on single image super-resolution: Dataset and
  study.
\newblock In \emph{Proceedings of the IEEE conference on computer vision and
  pattern recognition workshops}, pp.\  126--135, 2017.

\bibitem[Aubry et~al.(2014)Aubry, Paris, Hasinoff, Kautz, and Durand]{LLF}
Aubry, M., Paris, S., Hasinoff, S.~W., Kautz, J., and Durand, F.
\newblock Fast local laplacian filters: Theory and applications.
\newblock \emph{ACM Transactions on Graphics (TOG)}, 33\penalty0 (5):\penalty0
  1--14, 2014.

\bibitem[Bar et~al.(2022)Bar, Gandelsman, Darrell, Globerson, and
  Efros]{MAEVQGAN}
Bar, A., Gandelsman, Y., Darrell, T., Globerson, A., and Efros, A.
\newblock Visual prompting via image inpainting.
\newblock \emph{Advances in Neural Information Processing Systems},
  35:\penalty0 25005--25017, 2022.

\bibitem[Bevilacqua et~al.(2012)Bevilacqua, Roumy, Guillemot, and
  Alberi-Morel]{Set5}
Bevilacqua, M., Roumy, A., Guillemot, C., and Alberi-Morel, M.~L.
\newblock Low-complexity single-image super-resolution based on nonnegative
  neighbor embedding.
\newblock 2012.

\bibitem[Brown et~al.(2020)Brown, Mann, Ryder, Subbiah, Kaplan, Dhariwal,
  Neelakantan, Shyam, Sastry, Askell, et~al.]{GPT3}
Brown, T., Mann, B., Ryder, N., Subbiah, M., Kaplan, J.~D., Dhariwal, P.,
  Neelakantan, A., Shyam, P., Sastry, G., Askell, A., et~al.
\newblock Language models are few-shot learners.
\newblock \emph{Advances in neural information processing systems},
  33:\penalty0 1877--1901, 2020.

\bibitem[Bychkovsky et~al.(2011)Bychkovsky, Paris, Chan, and Durand]{FiveK}
Bychkovsky, V., Paris, S., Chan, E., and Durand, F.
\newblock Learning photographic global tonal adjustment with a database of
  input/output image pairs.
\newblock In \emph{CVPR 2011}, pp.\  97--104. IEEE, 2011.

\bibitem[Chen et~al.(2023)Chen, Yao, Chen, Zhang, and
  Liu]{chen2023understanding}
Chen, A., Yao, Y., Chen, P.-Y., Zhang, Y., and Liu, S.
\newblock Understanding and improving visual prompting: A label-mapping
  perspective.
\newblock In \emph{Proceedings of the IEEE/CVF Conference on Computer Vision
  and Pattern Recognition}, pp.\  19133--19143, 2023.

\bibitem[Chen et~al.(2021)Chen, Wang, Guo, Xu, Deng, Liu, Ma, Xu, Xu, and
  Gao]{IPT}
Chen, H., Wang, Y., Guo, T., Xu, C., Deng, Y., Liu, Z., Ma, S., Xu, C., Xu, C.,
  and Gao, W.
\newblock Pre-trained image processing transformer.
\newblock In \emph{Proceedings of the IEEE/CVF Conference on Computer Vision
  and Pattern Recognition}, pp.\  12299--12310, 2021.

\bibitem[Deng et~al.(2009)Deng, Dong, Socher, Li, Li, and Fei-Fei]{ImageNet}
Deng, J., Dong, W., Socher, R., Li, L.-J., Li, K., and Fei-Fei, L.
\newblock Imagenet: A large-scale hierarchical image database.
\newblock In \emph{2009 IEEE conference on computer vision and pattern
  recognition}, pp.\  248--255. Ieee, 2009.

\bibitem[Dong et~al.(2015)Dong, Loy, He, and Tang]{SRCNN}
Dong, C., Loy, C.~C., He, K., and Tang, X.
\newblock Image super-resolution using deep convolutional networks.
\newblock \emph{IEEE transactions on pattern analysis and machine
  intelligence}, 38\penalty0 (2):\penalty0 295--307, 2015.

\bibitem[Dong et~al.(2016)Dong, Loy, and Tang]{FSRCNN}
Dong, C., Loy, C.~C., and Tang, X.
\newblock Accelerating the super-resolution convolutional neural network.
\newblock In \emph{Computer Vision--ECCV 2016: 14th European Conference,
  Amsterdam, The Netherlands, October 11-14, 2016, Proceedings, Part II 14},
  pp.\  391--407. Springer, 2016.

\bibitem[Dosovitskiy et~al.(2020)Dosovitskiy, Beyer, Kolesnikov, Weissenborn,
  Zhai, Unterthiner, Dehghani, Minderer, Heigold, Gelly, et~al.]{ViT}
Dosovitskiy, A., Beyer, L., Kolesnikov, A., Weissenborn, D., Zhai, X.,
  Unterthiner, T., Dehghani, M., Minderer, M., Heigold, G., Gelly, S., et~al.
\newblock An image is worth 16x16 words: Transformers for image recognition at
  scale.
\newblock \emph{arXiv preprint arXiv:2010.11929}, 2020.

\bibitem[He et~al.(2022)He, Chen, Xie, Li, Doll{\'a}r, and Girshick]{MAE}
He, K., Chen, X., Xie, S., Li, Y., Doll{\'a}r, P., and Girshick, R.
\newblock Masked autoencoders are scalable vision learners.
\newblock In \emph{Proceedings of the IEEE/CVF conference on computer vision
  and pattern recognition}, pp.\  16000--16009, 2022.

\bibitem[Huang et~al.(2015)Huang, Singh, and Ahuja]{Urban100}
Huang, J.-B., Singh, A., and Ahuja, N.
\newblock Single image super-resolution from transformed self-exemplars.
\newblock In \emph{Proceedings of the IEEE conference on computer vision and
  pattern recognition}, pp.\  5197--5206, 2015.

\bibitem[Kirillov et~al.(2023)Kirillov, Mintun, Ravi, Mao, Rolland, Gustafson,
  Xiao, Whitehead, Berg, Lo, et~al.]{SAM}
Kirillov, A., Mintun, E., Ravi, N., Mao, H., Rolland, C., Gustafson, L., Xiao,
  T., Whitehead, S., Berg, A.~C., Lo, W.-Y., et~al.
\newblock Segment anything.
\newblock \emph{arXiv preprint arXiv:2304.02643}, 2023.

\bibitem[Kupyn et~al.(2018)Kupyn, Budzan, Mykhailych, Mishkin, and
  Matas]{DeblurGAN}
Kupyn, O., Budzan, V., Mykhailych, M., Mishkin, D., and Matas, J.
\newblock Deblurgan: Blind motion deblurring using conditional adversarial
  networks.
\newblock In \emph{Proceedings of the IEEE conference on computer vision and
  pattern recognition}, pp.\  8183--8192, 2018.

\bibitem[Li et~al.(2018)Li, Ren, Fu, Tao, Feng, Zeng, and Wang]{RESIDE}
Li, B., Ren, W., Fu, D., Tao, D., Feng, D., Zeng, W., and Wang, Z.
\newblock Benchmarking single-image dehazing and beyond.
\newblock \emph{IEEE Transactions on Image Processing}, 28\penalty0
  (1):\penalty0 492--505, 2018.

\bibitem[Li et~al.(2022)Li, Liu, Hu, Wu, Lv, and Peng]{AirNet}
Li, B., Liu, X., Hu, P., Wu, Z., Lv, J., and Peng, X.
\newblock All-in-one image restoration for unknown corruption.
\newblock In \emph{Proceedings of the IEEE/CVF Conference on Computer Vision
  and Pattern Recognition}, pp.\  17452--17462, 2022.

\bibitem[Liang et~al.(2021)Liang, Cao, Sun, Zhang, Van~Gool, and
  Timofte]{SwinIR}
Liang, J., Cao, J., Sun, G., Zhang, K., Van~Gool, L., and Timofte, R.
\newblock Swinir: Image restoration using swin transformer.
\newblock In \emph{Proceedings of the IEEE/CVF International Conference on
  Computer Vision}, pp.\  1833--1844, 2021.

\bibitem[Liu et~al.(2021)Liu, Shen, Zhang, Dolan, Carin, and
  Chen]{liu2021makes}
Liu, J., Shen, D., Zhang, Y., Dolan, B., Carin, L., and Chen, W.
\newblock What makes good in-context examples for gpt-$3 $?
\newblock \emph{arXiv preprint arXiv:2101.06804}, 2021.

\bibitem[Liu et~al.(2023)Liu, Yuan, Fu, Jiang, Hayashi, and Neubig]{liu2023pre}
Liu, P., Yuan, W., Fu, J., Jiang, Z., Hayashi, H., and Neubig, G.
\newblock Pre-train, prompt, and predict: A systematic survey of prompting
  methods in natural language processing.
\newblock \emph{ACM Computing Surveys}, 55\penalty0 (9):\penalty0 1--35, 2023.

\bibitem[Liu et~al.(2022)Liu, He, Chen, Zhang, Zhao, Dong, and Qiao]{CSRNet}
Liu, Y., He, J., Chen, X., Zhang, Z., Zhao, H., Dong, C., and Qiao, Y.
\newblock Very lightweight photo retouching network with conditional sequential
  modulation.
\newblock \emph{IEEE Transactions on Multimedia}, 2022.

\bibitem[Loshchilov \& Hutter(2017)Loshchilov and Hutter]{AdamW}
Loshchilov, I. and Hutter, F.
\newblock Decoupled weight decay regularization.
\newblock \emph{arXiv preprint arXiv:1711.05101}, 2017.

\bibitem[Ma et~al.(2023)Ma, Cheng, Wang, Zhang, Wang, and Zhang]{ProRes}
Ma, J., Cheng, T., Wang, G., Zhang, Q., Wang, X., and Zhang, L.
\newblock Prores: Exploring degradation-aware visual prompt for universal image
  restoration.
\newblock \emph{arXiv preprint arXiv:2306.13653}, 2023.

\bibitem[Martin et~al.(2001)Martin, Fowlkes, Tal, and Malik]{BSDS100}
Martin, D., Fowlkes, C., Tal, D., and Malik, J.
\newblock A database of human segmented natural images and its application to
  evaluating segmentation algorithms and measuring ecological statistics.
\newblock In \emph{Proceedings Eighth IEEE International Conference on Computer
  Vision. ICCV 2001}, volume~2, pp.\  416--423. IEEE, 2001.

\bibitem[Matsui et~al.(2017)Matsui, Ito, Aramaki, Fujimoto, Ogawa, Yamasaki,
  and Aizawa]{Manga109}
Matsui, Y., Ito, K., Aramaki, Y., Fujimoto, A., Ogawa, T., Yamasaki, T., and
  Aizawa, K.
\newblock Sketch-based manga retrieval using manga109 dataset.
\newblock \emph{Multimedia Tools and Applications}, 76\penalty0 (20):\penalty0
  21811--21838, 2017.

\bibitem[Min et~al.(2022)Min, Lyu, Holtzman, Artetxe, Lewis, Hajishirzi, and
  Zettlemoyer]{min2022rethinking}
Min, S., Lyu, X., Holtzman, A., Artetxe, M., Lewis, M., Hajishirzi, H., and
  Zettlemoyer, L.
\newblock Rethinking the role of demonstrations: What makes in-context learning
  work?
\newblock \emph{arXiv preprint arXiv:2202.12837}, 2022.

\bibitem[Potlapalli et~al.(2023)Potlapalli, Zamir, Khan, and Khan]{PromptIR}
Potlapalli, V., Zamir, S.~W., Khan, S., and Khan, F.~S.
\newblock Promptir: Prompting for all-in-one blind image restoration.
\newblock \emph{arXiv preprint arXiv:2306.13090}, 2023.

\bibitem[Radford et~al.(2019)Radford, Wu, Child, Luan, Amodei, Sutskever,
  et~al.]{GPT2}
Radford, A., Wu, J., Child, R., Luan, D., Amodei, D., Sutskever, I., et~al.
\newblock Language models are unsupervised multitask learners.
\newblock \emph{OpenAI blog}, 1\penalty0 (8):\penalty0 9, 2019.

\bibitem[Richardson(1972)]{richardson1972bayesian}
Richardson, W.~H.
\newblock Bayesian-based iterative method of image restoration.
\newblock \emph{JoSA}, 62\penalty0 (1):\penalty0 55--59, 1972.

\bibitem[Sun et~al.(2023)Sun, Chen, Wang, Wang, and Li]{sun2023exploring}
Sun, Y., Chen, Q., Wang, J., Wang, J., and Li, Z.
\newblock Exploring effective factors for improving visual in-context learning.
\newblock \emph{arXiv preprint arXiv:2304.04748}, 2023.

\bibitem[Wang et~al.(2021{\natexlab{a}})Wang, Wang, Dong, Xu, Yang, An, and
  Guo]{DASR}
Wang, L., Wang, Y., Dong, X., Xu, Q., Yang, J., An, W., and Guo, Y.
\newblock Unsupervised degradation representation learning for blind
  super-resolution.
\newblock In \emph{Proceedings of the IEEE/CVF Conference on Computer Vision
  and Pattern Recognition}, pp.\  10581--10590, 2021{\natexlab{a}}.

\bibitem[Wang et~al.(2023{\natexlab{a}})Wang, Dai, Chen, Huang, Li, Zhu, Hu,
  Lu, Lu, Li, et~al.]{InternImage}
Wang, W., Dai, J., Chen, Z., Huang, Z., Li, Z., Zhu, X., Hu, X., Lu, T., Lu,
  L., Li, H., et~al.
\newblock Internimage: Exploring large-scale vision foundation models with
  deformable convolutions.
\newblock In \emph{Proceedings of the IEEE/CVF Conference on Computer Vision
  and Pattern Recognition}, pp.\  14408--14419, 2023{\natexlab{a}}.

\bibitem[Wang et~al.(2021{\natexlab{b}})Wang, Xie, Dong, and Shan]{RealESRGAN}
Wang, X., Xie, L., Dong, C., and Shan, Y.
\newblock Real-esrgan: Training real-world blind super-resolution with pure
  synthetic data.
\newblock In \emph{Proceedings of the IEEE/CVF international conference on
  computer vision}, pp.\  1905--1914, 2021{\natexlab{b}}.

\bibitem[Wang et~al.(2023{\natexlab{b}})Wang, Wang, Cao, Shen, and
  Huang]{Painter}
Wang, X., Wang, W., Cao, Y., Shen, C., and Huang, T.
\newblock Images speak in images: A generalist painter for in-context visual
  learning.
\newblock In \emph{Proceedings of the IEEE/CVF Conference on Computer Vision
  and Pattern Recognition}, pp.\  6830--6839, 2023{\natexlab{b}}.

\bibitem[Wang et~al.(2022{\natexlab{a}})Wang, Li, Li, He, Huang, Zhao, Zhang,
  Xu, Liu, Wang, et~al.]{InternVideo}
Wang, Y., Li, K., Li, Y., He, Y., Huang, B., Zhao, Z., Zhang, H., Xu, J., Liu,
  Y., Wang, Z., et~al.
\newblock Internvideo: General video foundation models via generative and
  discriminative learning.
\newblock \emph{arXiv preprint arXiv:2212.03191}, 2022{\natexlab{a}}.

\bibitem[Wang et~al.(2022{\natexlab{b}})Wang, Cun, Bao, Zhou, Liu, and
  Li]{Uformer}
Wang, Z., Cun, X., Bao, J., Zhou, W., Liu, J., and Li, H.
\newblock Uformer: A general u-shaped transformer for image restoration.
\newblock In \emph{Proceedings of the IEEE/CVF Conference on Computer Vision
  and Pattern Recognition}, pp.\  17683--17693, 2022{\natexlab{b}}.

\bibitem[Wei et~al.(2018)Wei, Wang, Yang, and Liu]{LOL}
Wei, C., Wang, W., Yang, W., and Liu, J.
\newblock Deep retinex decomposition for low-light enhancement.
\newblock \emph{arXiv preprint arXiv:1808.04560}, 2018.

\bibitem[Wei et~al.(2023)Wei, Wei, Tay, Tran, Webson, Lu, Chen, Liu, Huang,
  Zhou, et~al.]{wei2023larger}
Wei, J., Wei, J., Tay, Y., Tran, D., Webson, A., Lu, Y., Chen, X., Liu, H.,
  Huang, D., Zhou, D., et~al.
\newblock Larger language models do in-context learning differently.
\newblock \emph{arXiv preprint arXiv:2303.03846}, 2023.

\bibitem[Xu et~al.(2011)Xu, Lu, Xu, and Jia]{xu2011image}
Xu, L., Lu, C., Xu, Y., and Jia, J.
\newblock Image smoothing via l 0 gradient minimization.
\newblock In \emph{Proceedings of the 2011 SIGGRAPH Asia conference}, pp.\
  1--12, 2011.

\bibitem[Yang et~al.(2023)Yang, Gao, Li, Gao, Wang, and Zheng]{TAM}
Yang, J., Gao, M., Li, Z., Gao, S., Wang, F., and Zheng, F.
\newblock Track anything: Segment anything meets videos.
\newblock \emph{arXiv preprint arXiv:2304.11968}, 2023.

\bibitem[Yu et~al.(2023)Yu, Feng, Feng, Liu, Jin, Zeng, and Chen]{IAM}
Yu, T., Feng, R., Feng, R., Liu, J., Jin, X., Zeng, W., and Chen, Z.
\newblock Inpaint anything: Segment anything meets image inpainting.
\newblock \emph{arXiv preprint arXiv:2304.06790}, 2023.

\bibitem[Zamir et~al.(2021)Zamir, Arora, Khan, Hayat, Khan, Yang, and
  Shao]{MPRNet}
Zamir, S.~W., Arora, A., Khan, S., Hayat, M., Khan, F.~S., Yang, M.-H., and
  Shao, L.
\newblock Multi-stage progressive image restoration.
\newblock In \emph{Proceedings of the IEEE/CVF conference on computer vision
  and pattern recognition}, pp.\  14821--14831, 2021.

\bibitem[Zamir et~al.(2022)Zamir, Arora, Khan, Hayat, Khan, and
  Yang]{Restormer}
Zamir, S.~W., Arora, A., Khan, S., Hayat, M., Khan, F.~S., and Yang, M.-H.
\newblock Restormer: Efficient transformer for high-resolution image
  restoration.
\newblock In \emph{Proceedings of the IEEE/CVF Conference on Computer Vision
  and Pattern Recognition}, pp.\  5728--5739, 2022.

\bibitem[Zeyde et~al.(2010)Zeyde, Elad, and Protter]{Set14}
Zeyde, R., Elad, M., and Protter, M.
\newblock On single image scale-up using sparse-representations.
\newblock In \emph{International conference on curves and surfaces}, pp.\
  711--730. Springer, 2010.

\bibitem[Zhang et~al.(2017)Zhang, Zuo, Chen, Meng, and Zhang]{DNCNN}
Zhang, K., Zuo, W., Chen, Y., Meng, D., and Zhang, L.
\newblock Beyond a gaussian denoiser: Residual learning of deep cnn for image
  denoising.
\newblock \emph{IEEE transactions on image processing}, 26\penalty0
  (7):\penalty0 3142--3155, 2017.

\bibitem[Zhang et~al.(2021)Zhang, Liang, Van~Gool, and Timofte]{BSRGAN}
Zhang, K., Liang, J., Van~Gool, L., and Timofte, R.
\newblock Designing a practical degradation model for deep blind image
  super-resolution.
\newblock In \emph{Proceedings of the IEEE/CVF International Conference on
  Computer Vision}, pp.\  4791--4800, 2021.

\bibitem[Zhang et~al.(2023)Zhang, Zhou, and Liu]{zhang2023makes}
Zhang, Y., Zhou, K., and Liu, Z.
\newblock What makes good examples for visual in-context learning?
\newblock \emph{arXiv preprint arXiv:2301.13670}, 2023.

\end{thebibliography}
	\bibliographystyle{icml2024}

	\newpage

	\appendix
	\onecolumn
	\section{Details of Image Processing Tasks}
	\subsection{Image Restoration Tasks}
	For image restoration, we consider the following 10 different degradation types.
	
	\noindent\textbf{Gaussian Noise.} Gaussian noise is a type of random variation that affects the pixel values of an image. It is characterized by a probability distribution known as the Gaussian distribution or normal distribution. In images, Gaussian noise appears as a random variation in pixel values, where the noise values follow the Gaussian distribution pattern. We add Gaussian noise to the clean images to synthesize noisy images. The noise level is uniformly sampled from [10, 50].
	
	\noindent\textbf{Gaussian Blur.} Gaussian blur is a common image filtering technique used to reduce high-frequency noise and details in an image. It employs a mathematical function known as the Gaussian kernel to apply weighted averaging to the pixel values within a specified neighborhood. We apply isotropic Gaussian blur kernel on the clean image to synthesize blurry images. The kernel width is uniformly sampled from [2, 4].
	
	\noindent\textbf{Poisson Noise.} Poisson noise is manifested as random variations in pixel intensities, leading to a grainy appearance. It follows a Poisson distribution, where the variance is proportional to the mean intensity of the signal. Following the common settings, we set the noise level to 2.
	
	\noindent\textbf{Salt \& Pepper Noise.} Salt and Pepper noise, also known as impulse noise, is a type of irregular interference commonly found in digital images. This noise is characterized by random occurrences of very bright (salt) and very dark (pepper) pixels. We simulate the salt and pepper noise with a signal noise ratio of 0.95. The probabilities of producing salt and pepper are both 50\%. 
	
	\noindent\textbf{Jpeg Compression Artifacts.} JPEG compression artifacts are distortions or anomalies that arise due to the lossy nature of the JPEG compression process. These artifacts typically manifest as blocky patterns, color shifts, and blurriness, especially in areas with high contrast or fine details. The visual quality factor is uniformly sampled from [10, 40].
	
	\noindent\textbf{Ringing Artifacts.} Ringing artifacts are most noticeable as a pair of bright and dark bands adjacent to the edges of objects in the image. These bands result from the reconstruction process during compression or other image transformations. When a sharp edge is compressed or enhanced, the algorithm may introduce extra pixel values that weren't present in the original image. This results in an overemphasis of the edge and the creation of the distinct bright and dark bands. We employ the implementation of ring artifacts from Real-ESRGAN \cite{RealESRGAN}.

	\noindent\textbf{R-L Algorithm.} The Richardson-Lucy (R-L) algorithm aims to estimate the original, sharp image from a degraded version by iteratively updating the estimate based on the observed degraded image and a point spread function (PSF) that characterizes the blurring. It iteratively refines the estimated image to minimize the difference between the observed and estimated images, while accounting for the effects of blurring and noise. We directly employ python built-in function \texttt{skimage.restoration.richardson\_lucy} to attain the processed results.
	
	\noindent\textbf{Inpainting.} We randomly add masked streaks in the clean image, obtaining masked images to be fulfilled. The number of streaks ranges from 5 to 10 and the thickness is sampled from [5, 10].
	
	\noindent\textbf{Haze.} For image dehazing, we utilize the ITS training set of RESIDE dataset \cite{RESIDE} for training and the SOTS-indoor dataset for testing.
	
	\noindent\textbf{Rain.} We employ two types of rain addition models: Simple Rain Model and Complex Rain Model. The Simple Rain Model is a straightforward additive rain model, and we directly synthesize it on the clean images. The Complex Rain Model utilizes Rain13K dataset \cite{MPRNet}, including an assortment of diverse rain models. Test100 dataset is adopted for evaluation.
	
	\subsection{Image Enhancement Tasks}
	\noindent\textbf{Low-light Image Enhancement.} The goal of low-light image enhancement is to adjust the image's brightness, contrast, and color balance while preserving important details and minimizing noise. We adopt the commonly-used LOL dataset \cite{LOL} for training and testing.

	\noindent\textbf{Local Laplacian Filtering.} In local Laplacian filtering \cite{LLF}, a Laplacian pyramid is constructed for the input image, representing different scales of details. This pyramid is modified using a control grid that adjusts the appearance of image regions based on their contrast and brightness levels. By applying different filters to the pyramid's levels, the method enhances the image's finer details while respecting its global structure. Following \cite{CSRNet}, we apply this operator on images retouched by expert C of the MIT-Adobe FiveK dataset \cite{FiveK}. The training and testing sets of MIT-Adobe FiveK dataset are adopted.
	
	\subsection{Image Edge Detection}
	\noindent\textbf{Canny Operator.} The Canny edge detection operator is a widely used method in image processing for detecting edges in digital images. It aims to identify significant changes in intensity within an image, which often correspond to object boundaries or important features. The Canny operator is known for its ability to detect edges accurately, suppressing noise and responding well to significant changes in intensity. We set the \texttt{threshold 1} and the \texttt{threshold 2} of Canny operator as 50 and 200.
	
	\noindent\textbf{Laplacian Operator.} The Laplacian operator is a mathematical filter commonly used for detecting regions of rapid intensity changes. Mathematically, the Laplacian operator calculates the second derivative of the image intensity with respect to its spatial coordinates (x and y).
	
	\section{Limitations and Prospectives}
	PromptGIP exhibits considerable potential in addressing a diverse range of image processing tasks through its innovative visual prompting question-answering paradigm. However, there are certain limitations and areas for further exploration that merit attention. These limitations and potential areas for improvement could provide valuable insights for future research and development efforts.
	
	PromptGIP excels at tasks guided by explicit prompts, but its current scope does not extend to generating unexpected or emergent outcomes. Our findings suggest that the model still lacks the ability to generate novel solutions beyond what it has learned from training data. This observation is consistent with previous investigations of language models~\cite{min2022rethinking}. Recent works demonstrate that the effectiveness of in-context learning for large language models heavily relies on the quality, diversity, and quantity of the training data~\cite{wei2023larger}. Inadequate or biased training data can lead to suboptimal performance on certain tasks or scenarios. Due to constrained computational resources and limited training data, we are currently unable to conduct extensive large scaling-up experiments. However, we believe that in-context learning still has the potential to show more impressive effects.
	
	Another limitation lies in the current backbone choice of ViT. ViT \cite{ViT} splits the input image into 16$\times$16 patches and transforms them into a sequence of linear embeddings. The rough patch-splitting strategy would cause a significant loss of high-frequency information, such as edges, textures and structures, leading to severe artifacts and over-smoothed results. Consequently, current Transformer-based low-level models \cite{Restormer,Uformer,SwinIR,IPT} still adopt CNN for pre/post-processing. In this paper, we mainly focus on validating the effectiveness of prompt learning in multitask image processing, thus we simply adopt the basic ViT architecure to conduct experiments. However, the original ViT backbone cannot achieve supreme performance on low-level vision tasks, leading to subpar quantitative results compared to state-of-the art Restormer backbone. This issue could be addressed by adopting stronger backbone models in the future work.
	
	\begin{figure*}[htbp]
		\centering
		\includegraphics[width=0.8\linewidth]{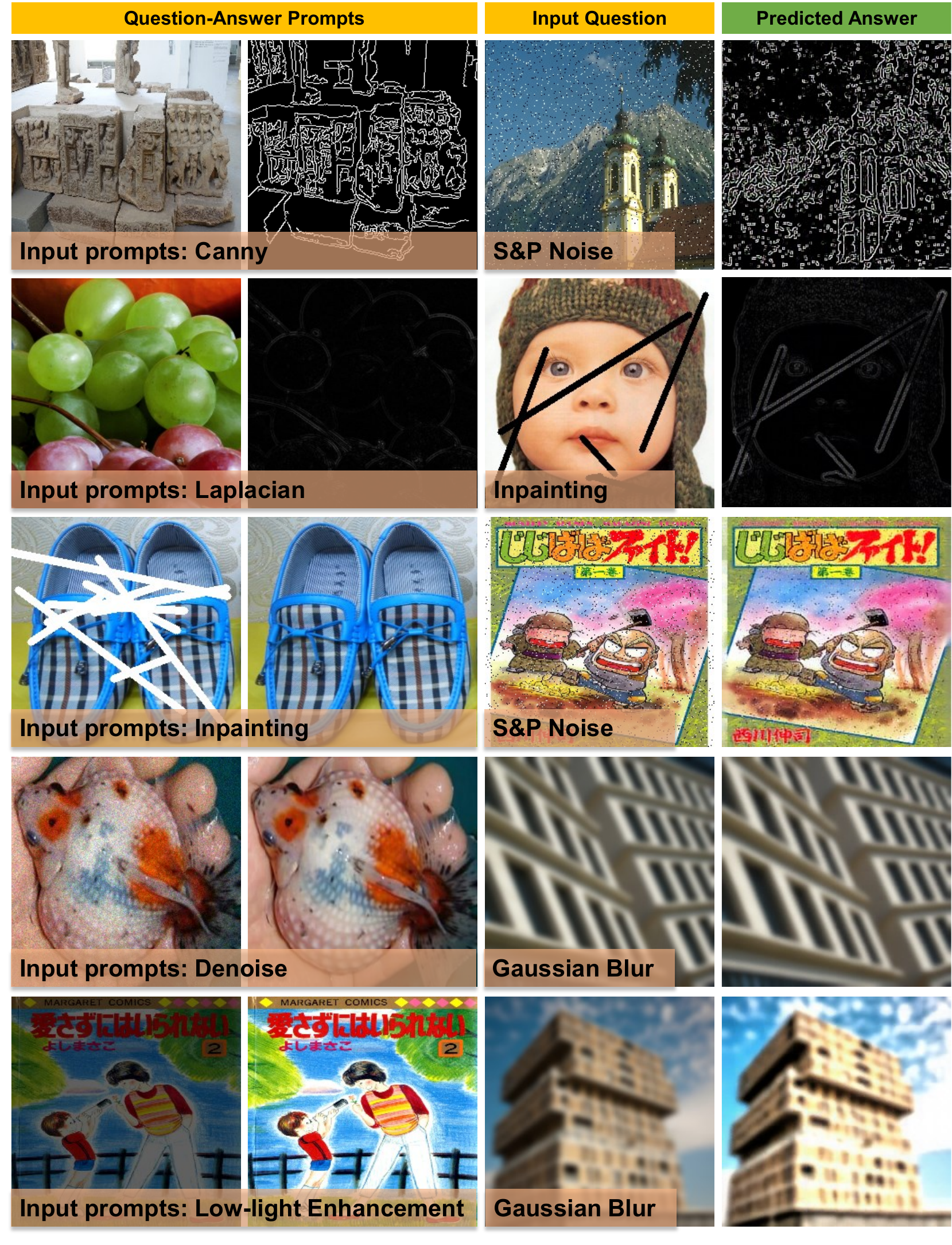} 
		\caption{Effectiveness of visual prompts in PromptGIP. PromptGIP can correctly identity and execute the designed tasks, instead of memorizing the types of input image.}
		\label{fig:prompt_test1}
	\end{figure*}
	
	\section{Effects of Visual Prompts}
	\subsection{Validating the Efficacy of Visual Prompts in PromptGIP}
	As discussed in the main paper, we find that the utilization of prompts in Painter does not yield the anticipated outcomes. Instead, the model tends to exhibit suboptimal behavior by overly adapting to specific tasks and the corresponding datasets. Specifically, when presented with a noisy image pair, the model is intended to perform denoising on the input query image. However, a noteworthy observation is that when the input query image portrays an indoor bedroom scene, the model erroneously engages in depth estimation rather than denoising. This behavior arises due to the fact that the training data for depth estimation tasks predominantly consist of indoor scene images. Consequently, the model associates depth estimation tasks with indoor images. As a result, when the input query image resembles an indoor setting, the model instinctively conducts depth estimation processing. Consequently, it becomes evident that the utilization of visual prompts in Painter does not yield the intended results.

	To assess the potency of visual prompts within the PromptGIP framework, we conducted a series of tests. Illustrated in Fig. \ref{fig:prompt_test1}, our experiments span different scenarios. When tasked with a canny operator prompt, PromptGIP adeptly applies the Canny operator to input query images afflicted with salt and pepper noise. Notably, the model's performance is not driven by data memorization for denoising operations based on the query image content. Further evaluations reveal intriguing dynamics. In instances where the task prompt involves denoising and the query image exhibits Gaussian blur, PromptGIP appropriately refrains from processing the image. This behavior underscores the model's grasp of the task prompt's intent, resisting unnecessary processing when incongruities are detected. Moreover, when presented with a pair of low-light and high-light images in a question-answer format, PromptGIP deftly heightens the brightness of the query image. It is intriguing to observe that this occurs even when the input query image bears blurriness. Notably, the model abstains from undertaking deblurring actions, signaling its grasp of the task prompt's essence. These comprehensive assessments affirm that PromptGIP possesses the ability to comprehend and act upon task prompts without succumbing to the perils of overfitting to specific training data.

	\begin{figure*}[h]
		\centering
		\includegraphics[width=0.95\linewidth]{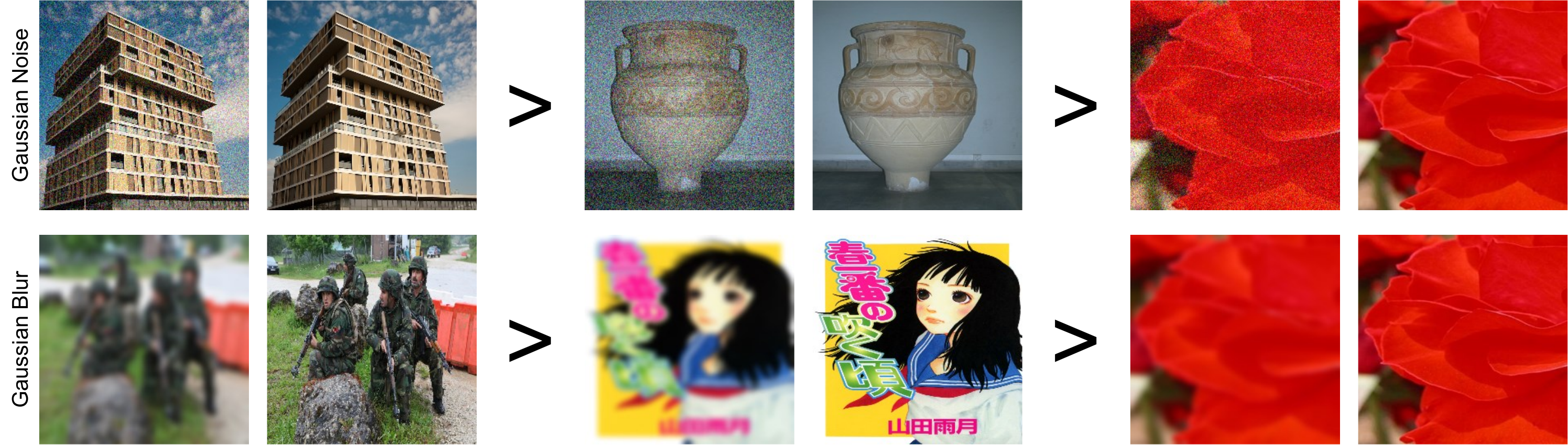} 
		\caption{Different visual prompts will lead to different results. The figure above ranks the effects of different visual prompts on Gaussian denoising and Gaussian deblurring tasks.}
		\label{fig:prompt_compare}
	\end{figure*}
	
	\begin{table}[htbp]
		\centering
		\begin{minipage}{.43\linewidth}
			\centering
			\setlength\tabcolsep{16pt}
			\renewcommand{\arraystretch}{1}
			\begin{tabular}{|c|c|c|}
				\hline
				Prompt ID & PSNR    & SSIM   \\ \hline
				Idx 0          & 26.1142 & 0.8131 \\ \hline
				Idx 1          & 25.9345 & 0.8043 \\ \hline
				Idx 2          & 25.9343 & 0.8097 \\ \hline
				Idx 3          & 26.1386 & 0.8127 \\ \hline
				Idx 4          & 25.8980 & 0.8099 \\ \hline
				Idx 5          & 25.8063 & 0.8101 \\ \hline
				Idx 6          & 26.0166 & 0.8102 \\ \hline
				Idx 7          & 25.9953 & 0.8087 \\ \hline
				Idx 8          & 25.9991 & 0.8121 \\ \hline
				Idx 9          & \textbf{26.2194} & \textbf{0.8167} \\ \hline
				Idx 10         & 26.0864 & 0.8105 \\ \hline
				Idx 11         & 26.0188 & 0.8110 \\ \hline
				Idx 12         & 26.0002 & 0.8115 \\ \hline
				Idx 13         & 26.0299 & 0.8143 \\ \hline
				Idx 14         & 26.0116 & 0.8129 \\ \hline
				Idx 15         & 25.6136 & 0.8048 \\ \hline
				Idx 16         & 26.1333 & 0.8124 \\ \hline
				Idx 17         & 25.8457 & 0.7930 \\ \hline
				Idx 18         & 23.2508 & 0.7458 \\ \hline
				Idx 19         & 26.0260 & 0.8061 \\ \hline
				Avg.           & 25.8536 & 0.8064 \\ \hline
				Std.           & 0.6110  & 0.0147 \\ \hline
			\end{tabular}
			\caption{Influence of employing different visual prompts on Gaussian denoise task.}
			\label{tab:GaussionNoise_prompts}
		\end{minipage}\hspace{0.8cm}%
		\begin{minipage}{.43\linewidth}
			\centering
			\setlength\tabcolsep{16pt}
			\renewcommand{\arraystretch}{1}
			\begin{tabular}{|c|c|c|}
				\hline
				Prompt ID & PSNR     & SSIM   \\ \hline
				Idx 0     & 22.0729  & 0.6280 \\ \hline
				Idx 1     & 22.7198  & 0.6901 \\ \hline
				Idx 2     & 22.4821  & 0.6810 \\ \hline
				Idx 3     & 22.4858  & 0.6894 \\ \hline
				Idx 4     & 21.6575  & 0.6089 \\ \hline
				Idx 5     & 22.5316  & 0.6726 \\ \hline
				Idx 6     & 22.6002  & 0.6909 \\ \hline
				Idx 7     & 22.6652  & 0.6779 \\ \hline
				Idx 8     & 22.6273  & 0.6728 \\ \hline
				Idx 9     & 22.7522  & 0.6866 \\ \hline
				Idx 10    & 21.9502  & 0.6826 \\ \hline
				Idx 11    & 22.7247  & 0.6793 \\ \hline
				Idx 12    & 22.7384  & 0.6809 \\ \hline
				Idx 13    & 22.4326  & 0.6861 \\ \hline
				Idx 14    & 22.4880  & 0.6803 \\ \hline
				Idx 15    & 22.5095  & 0.6802 \\ \hline
				Idx 16    & 22.7451  & 0.6810 \\ \hline
				Idx 17    & \textbf{22.7658}  & \textbf{0.6911} \\ \hline
				Idx 18    & 20.8271  & 0.6286 \\ \hline
				Idx 19    & 22.6869  & 0.6851 \\ \hline
				Avg.      & 22.4231  & 0.6737 \\ \hline
				Std.      & 0.4647   & 0.0226 \\ \hline
			\end{tabular}
			\caption{Influence of employing different visual prompts on Gaussian deblur task.}
			\label{tab:GaussionBlur_prompts}
		\end{minipage}
	\end{table}
	
	\subsection{Importance of Prompt Quality in Prompt Learning}
	The success of prompt learning hinges upon the quality and relevance of the provided prompts. The impact of ambiguous or poorly defined prompts on the final results underscores the critical role of prompt selection in this paradigm. In this context, our main paper showcases a meticulous process of prompt selection, where we curate 20 distinct prompts for each task, subsequently reporting the most favorable quantitative outcomes. The detailed outcomes of these prompt variations across several representative tasks are outlined in Tab. \ref{tab:GaussionNoise_prompts} and Tab. \ref{tab:GaussionBlur_prompts}.
	
	This exploration of prompt quality aligns with similar investigations conducted in fields like NLP and high-level vision \cite{liu2021makes,sun2023exploring,chen2023understanding,zhang2023makes}. Building upon this, we delve into a comparative analysis of visual prompts' impact on Gaussian denoising and Gaussian deblurring tasks in Fig. \ref{fig:prompt_compare}. Notably, the analysis underscores that the effectiveness of a visual prompt is closely tied to its richness in texture and color. Those prompts that embody these qualities consistently produce superior outcomes, thereby significantly enhancing overall model performance. The strategic crafting of prompts thus emerges as a pivotal determinant in unlocking the full potential of models such as ours.

	\section{More Visual Results}
\begin{figure*}[htbp]
		\centering
		\includegraphics[width=0.95\linewidth]{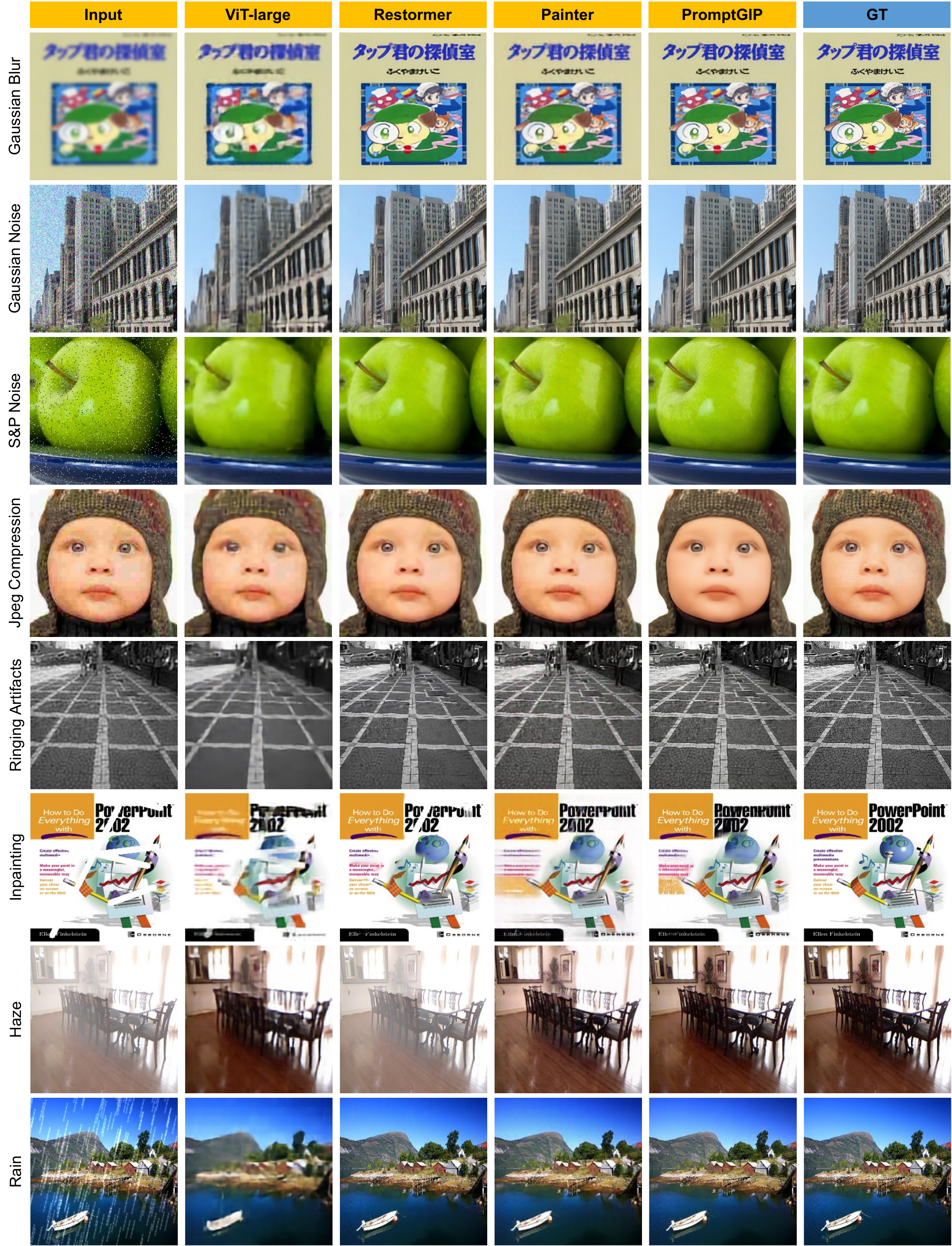} 
		\caption{Visual comparison on all-in-one image restoration task.}
		\label{fig:visual_compare1}
	\end{figure*}

\begin{figure*}[htbp]
		\centering
		\includegraphics[width=0.95\linewidth]{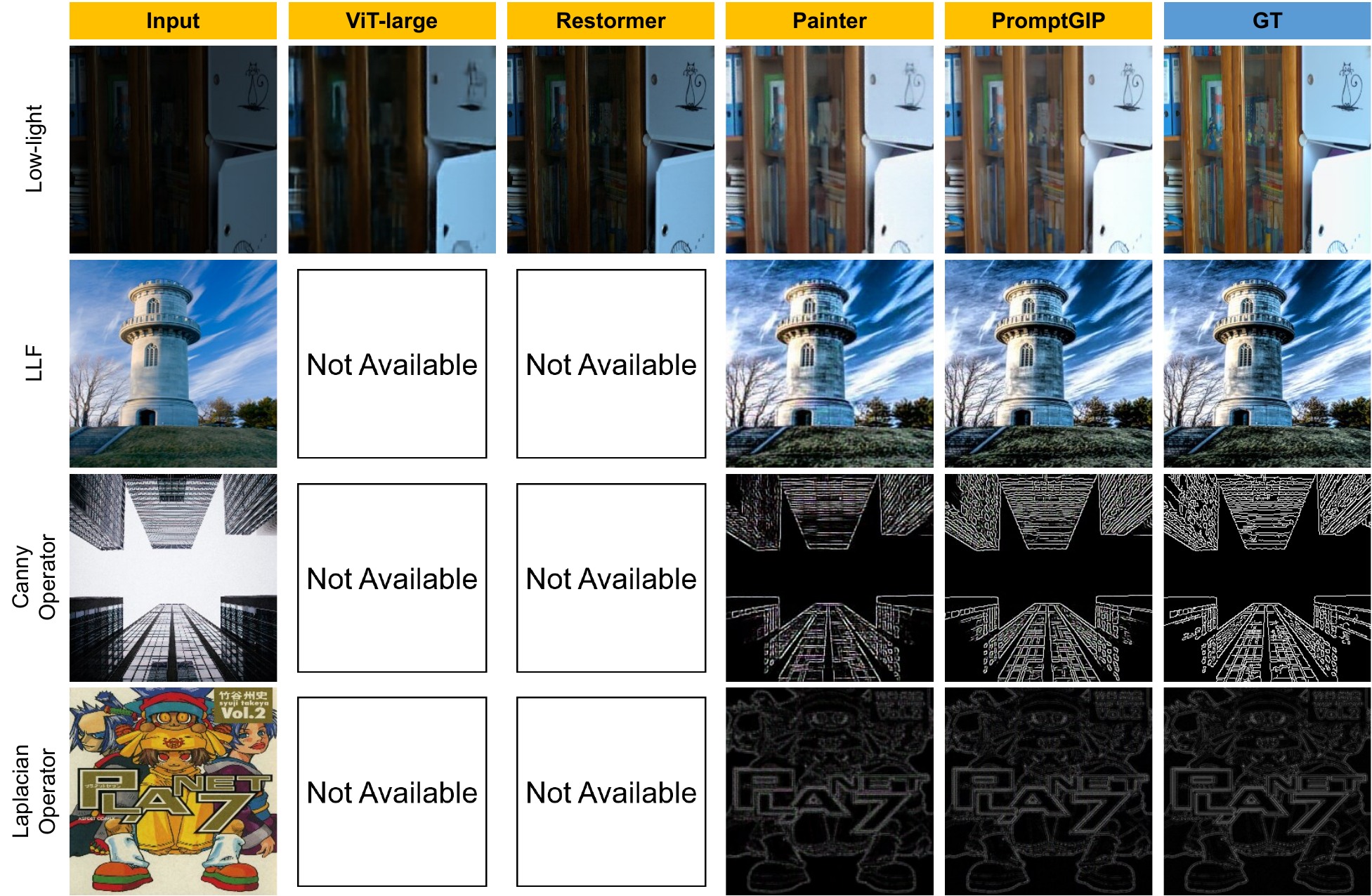} 
		\caption{Visual comparison on image enhancement and image edge detection tasks.}
		\label{fig:visual_compare2}
	\end{figure*}

	In order to comprehensively assess the performance of PromptGIP, we present an array of qualitative visual results across distinct tasks including image restoration, image enhancement, and image edge detection. In conjunction with this, a comparative evaluation is conducted against the well-established ViT-large model \cite{ViT} and the Restormer model \cite{Restormer}. The visual outputs and comparisons are thoughtfully illustrated in Fig. \ref{fig:visual_compare1} and \ref{fig:visual_compare2}.
	
	PromptGIP's proficiency in generating visually appealing outputs is readily evident in the presented results. Notably, the visual quality not only surpasses that of the baseline ViT model but also stands in competitive parity with the Restormer model. However, the significance of PromptGIP's capability extends beyond visual quality. The distinctive strength of PromptGIP becomes evident in its ability to effectively handle an extensive array of image enhancement tasks and image detection. This becomes a noteworthy distinction from traditional models, which often struggle to concurrently address such a diverse spectrum of tasks.

\end{document}